\documentclass[10pt,journal,compsoc]{IEEEtran}

\ifCLASSOPTIONcompsoc
\usepackage[nocompress]{cite}
\else
\usepackage{cite}
\fi

\usepackage{bbding}
\usepackage{times}
\usepackage{epsfig}
\usepackage{graphicx}
\usepackage{caption}
\graphicspath{{figures/}}
\usepackage{amsmath}
\usepackage{amssymb}
\usepackage[colorlinks,linkcolor=red]{hyperref}

\usepackage[utf8]{inputenc} 
\usepackage[T1]{fontenc}    
\usepackage{url}            
\usepackage{booktabs}       
\usepackage{amsfonts}       
\usepackage{nicefrac}       
\usepackage{microtype}      
\usepackage{algorithm}
\usepackage{algpseudocode}
\usepackage{amsfonts,amssymb}
\usepackage{array}
\usepackage{bm}
\usepackage{multirow}

\usepackage{times} 
\usepackage{helvet} 
\usepackage{courier} 
\usepackage{amsmath}
\usepackage{footnote}
\usepackage{multirow}
\usepackage{array}
\usepackage{algorithm}
\usepackage{algpseudocode}
\usepackage{amsfonts,amssymb}
\usepackage{booktabs}
\usepackage{xcolor}
\usepackage{verbatim}

\usepackage{ragged2e}

\usepackage{colortbl}
\usepackage{arydshln}
\usepackage{multirow}
\usepackage{multicol}
\usepackage{dashrule}
\usepackage{fancyhdr}

\definecolor{mygray}{gray}{.9}

\begin{document}

	\title{TransZero++: Cross Attribute-Guided Transformer for Zero-Shot Learning}

	\author{
		Shiming~Chen,
		Ziming~Hong,
		Wenjin~Hou,
		Guo-Sen~Xie,
		Yibing Song,
		Jian Zhao,
		Xinge~You,~\IEEEmembership{Senior~Member,~IEEE},
		Shuicheng~Yan,~\IEEEmembership{Fellow,~IEEE},
		and~Ling~Shao,~\IEEEmembership{Fellow,~IEEE}
		
		\thanks{S. Chen, Z. Hong, W. Hou and X. You are with the School of Electronic Information and Communication, Huazhong University of Science and Technology, Wuhan 430074, China. (Corresponding author: Xinge You. e-mail: youxg@hust.edu.cn)}
		\thanks{G.-S. Xie is with the Nanjing University of Science and Technology, China.}
		\thanks{Y. Song is with AI$^3$ Institute, Fudan University, Shanghai, China.}
		\thanks{J. Zhao is with the Institute of North Electronic Equipment, Beijing, China, and the Peng Cheng Laboratory, Shenzhen, China.}
		\thanks{S. Yan is with Sea AI Lab (SAIL), Singapore.}
		\thanks{L. Shao is with Terminus Group, China.}
	}

	\markboth{IEEE TRANSACTIONS ON PATTERN ANALYSIS AND MACHINE INTELLIGENCE}%
	{Chen \MakeLowercase{\textit{et al.}}: Bare Demo of IEEEtran.cls for Computer Society Journals}

	\IEEEtitleabstractindextext{%
		
		\begin{abstract}
			\justifying Zero-shot learning (ZSL) tackles the novel class recognition problem by transferring semantic knowledge from seen classes to unseen ones. Semantic knowledge is typically represented by attribute descriptions shared between different classes, which act as strong priors for localizing object attributes that represent discriminative region features, enabling significant and sufficient  visual-semantic interaction for advancing ZSL. Existing attention-based models have struggled to learn inferior region features in a single image by solely using unidirectional attention, which ignore the transferable and discriminative attribute localization of visual features for representing the key semantic knowledge for effective knowledge transfer in ZSL. In this paper, we propose a cross attribute-guided Transformer network, termed TransZero++, to refine visual features and learn accurate attribute localization for key semantic knowledge representations in ZSL. Specifically, TransZero++ employs an attribute$\rightarrow$visual Transformer sub-net (AVT) and a visual$\rightarrow$attribute Transformer sub-net (VAT) to learn attribute-based visual features and visual-based attribute features, respectively. By further introducing feature-level and prediction-level semantical collaborative losses, the two attribute-guided transformers teach each other to learn semantic-augmented visual embeddings for key semantic knowledge representations via semantical collaborative learning. Finally, the semantic-augmented visual embeddings learned by AVT and VAT are fused to conduct desirable visual-semantic interaction cooperated with class semantic vectors for ZSL classification. Extensive experiments show that TransZero++ achieves the new state-of-the-art results on three golden ZSL benchmarks and on the large-scale
			ImageNet dataset. The project website is available at: \url{https://shiming-chen.github.io/TransZero-pp/TransZero-pp.html}.
		\end{abstract}
		
		\begin{IEEEkeywords}
			Zero-Shot Learning; Transformer; Attribute Localization; Semantic-Augmented Visual Embedding; Semantical Collaborative Learning.
	\end{IEEEkeywords}}

	\maketitle

	\IEEEdisplaynontitleabstractindextext
	
	\IEEEpeerreviewmaketitle

	\section{Introduction and Motivation}\label{sec:introduction}
	
	\IEEEPARstart{H}{uman} beings are capable of learning novel concepts based on prior experience without seeing them in advance. For example, given the clues that zebras appear like horses yet with black-and-white stripes of tigers, one can quickly recognize a zebra if he/she has seen horses and tigers before. Nevertheless, unlike humans, supervised machine learning  models can only classify samples belonging to the classes that have already appeared during the training phase, and they are not capable of handling samples from previously unseen categories. Motivated by this challenge, zero-shot learning (ZSL) was proposed to recognize new classes by exploiting the intrinsic semantic relatedness during learning  \cite{Larochelle2008ZerodataLO,Palatucci2009ZeroshotLW,Lampert2009LearningTD,Lampert2014AttributeBasedCF,Fu2018ZeroShotLO,FuYanwei2015TransductiveMZ}. Since ZSL is a foundational method of artificial intelligence, it is commonly used in tasks with wide real-world applications, \textit{e.g.}, image classification \cite{Frome2013DeViSEAD,Xian2019FVAEGAND2AF}, image retrieval \cite{Shen2018ZeroShotSH, Dutta2020SemanticallyTP}, semantic segmentation \cite{Bucher2019ZeroShotSS} and object detection \cite{Bansal2018ZeroShotOD}. Particularly, the core idea of ZSL is to learn discriminative and transferable visual features for conducting effective visual-semantic interactions based on the semantic information (\textit{e.g.}, attribute descriptions \cite{Lampert2014AttributeBasedCF}, sentence embeddings \cite{Reed2016LearningDR}, and DNA \cite{Badirli2021FineGrainedZL}), which are shared between the seen and unseen classes employed to support the knowledge transfer.  At present, most existing ZSL methods bases on attribute descriptions. According to the different ranges of the label space during testing, ZSL methods can be categorized into conventional ZSL (CZSL), which aims to predict unseen classes, and generalized ZSL (GZSL), which can predict both seen and unseen classes  \cite{Xian2019ZeroShotLC}. Moreover, ZSL can also be classified as inductive ZSL \cite{Xian2018FeatureGN,Xie2021GeneralizedZL}, which only utilizes the labeled seen data, and transductive ZSL \cite{Song2018TransductiveUE,Xie2021vman}, assuming that unlabeled unseen data are available \cite{Xian2019ZeroShotLC}. Inductive ZSL is more reasonable and challenging, we are thus focused on the inductive ZSL setting in this paper. ZSL is typically denoted as zero-shot image classification or object recognition \cite{Lampert2009LearningTD,Lampert2014AttributeBasedCF}, which is different to zero-shot text classification \cite{Yin2019BenchmarkingZT} or zero-shot transfer of model \cite{Radford2021LearningTV}. We also follow this standard in this paper.

	\begin{figure*}[t]
		\centering
		\includegraphics[scale=0.33]{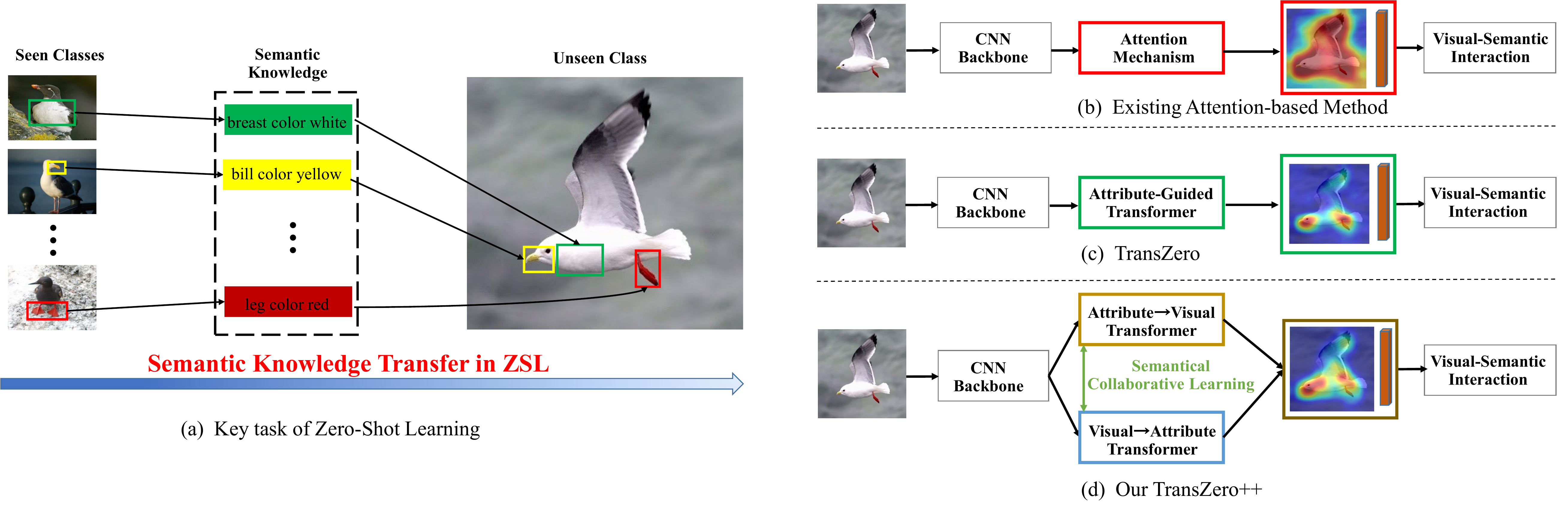}\vspace{-3mm}
		\caption{Motivation illustration. (a) ZSL discovers the discriminative and transferable semantic knowledge to enable efficient knowledge transfer from seen classes to unseen ones.  (b) Existing attention-based ZSL methods simply learn inferior region embeddings (e.g., the whole bird body) for semantic knowledge representations using unidirectional attention, ignoring the transferable and discriminative attribute localization (e.g., the distinctive bird body parts) of visual features; (c) TransZero \cite{Chen2021TransZero} employs an attribute-augmented Transformer to  reduce the entangled relationships among region features to improve their transferability, and localizes the object attributes to represent discriminative region features as important semantic knowledge. (d) Our TransZero++ takes two cross attribute-augmented Transformers (\textit{i.e.}, attribute$\rightarrow$visual Transformer and visual$\rightarrow$attribute Transformer) to further fully discover more intrinsic semantic knowledge via semantical collaborative learning, encouraging more desirable and sufficient visual-semantic interaction.}
		\label{fig:motivation}
	\end{figure*}
	
	To enable visual-semantic interactions for knowledge transfer from seen to unseen classes, early embedding-based ZSL methods \cite{Akata2016LabelEmbeddingFI,Song2018TransductiveUE,Li2018DiscriminativeLO} are trying to learn the embedding between seen classes and their class semantic vectors, and then classify unseen classes by nearest neighbor search in the embedding space. However, these embedding-based methods inevitably overfit to seen classes under the GZSL setting (known as the bias problem), since the embedding is only learned by seen class samples. To mitigate this bias problem, many generative ZSL methods have been proposed to synthesize feature samples for unseen classes by leveraging generative models (\textit{e.g.}, variational autoencoders (VAEs) \cite{Arora2018GeneralizedZL,Schnfeld2019GeneralizedZA,Chen2021HSVA}, generative adversarial nets (GANs) \cite{Xian2018FeatureGN,Xian2019FVAEGAND2AF,Chen2021FREE}, and generative flows \cite{Shen2020InvertibleZR}) for data augmentation. Thus the generative ZSL methods can compensate for the lack of training samples of unseen classes and convert ZSL into a supervised classification task.

	Although these methods have achieved significant improvement, they rely on global (holistic) visual features\footnote{These global visual features are directly extract from a CNN Backbone (\textit{e.g.}, ResNet \cite{He2016DeepRL}) pre-trained on ImageNet \cite{Deng2009ImageNetAL} alone, ignoring the cross-dataset bias between ImageNet and ZSL benchmarks (\textit{e.g.}, CUB \cite{Welinder2010CaltechUCSDB2}). Such a bias inevitably results in poor-quality visual features in which not all the dimensions are semantically related to the pre-defined attributes for ZSL tasks\cite{Chen2021FREE,Chen2021SemanticsDF}.} which are insufficient for capturing the fine-grained attribute information (\textit{e.g.}, “bill color yellow” of \textit{Red Legged Kittiwake
	}) for representing semantic knowledge \cite{Xie2020RegionGE,Huynh2020FineGrainedGZ,Wang2021RegionSA}. Because the discriminative and transferable semantic knowledge is usually contained in a few regions corresponding to a few attributes, enabling efficient knowledge transfer from seen classes to unseen ones, as shown in Fig. \ref{fig:motivation} (a). Thus, the visual feature representations learned by these methods are inferior, resulting in undesirable visual-semantic interactions for knowledge transfer. More recently, few attention-based models \cite{Xie2019AttentiveRE,Xie2020RegionGE,Zhu2019SemanticGuidedML,Xu2020AttributePN,Yu2018StackedSA,Liu2019AttributeAF,Hong2022SemanticCE} have attempted to explore more discriminative region features, as shown in Fig. \ref{fig:motivation} (b). However, these methods are limited in: i) they directly take the entangled region (grid) features\footnote{These entangled region (grid) features usually include relative geometry relationship priors between different regions\cite{Xie2020RegionGE}.} for ZSL classification, which hinders the transferability of visual features from seen to unseen classes; ii) they simply learn region embeddings (\textit{e.g.}, the whole bird body) using unidirectional attention, neglecting the importance of discriminative attribute localization (\textit{e.g.}, the distinctive bird body parts) for key semantic knowledge representations. Thus, properly discovering key semantic knowledge from visual features to enable efficient semantic knowledge transfer in ZSL has become very necessary.
	
	To tackle the above challenges, in this paper, we propose a cross attribute-guided Transformer, termed TransZero++, which discovers the key semantic knowledge for efficient knowledge transfer from seen classes to unseen ones in ZSL via semantical collaborative learning, as shown in Fig. \ref{fig:motivation} (d). Specifically, TransZero++ consists of two attribute-guided Transformer sub-nets  (\textit{i.e.}, attribute$\rightarrow$visual Transformer (AVT) and visual$\rightarrow$attribute Transformer (VAT)) that learn attribute-based visual features and visual-based attribute features respectively, which are further mapped into the semantic embedding space using two mapping functions $\mathcal{M}_1$ and $\mathcal{M}_2$ to conduct desirable visual-semantic interaction. In AVT and VAT, we first take a feature augmentation encoder to augment visual features by  i) alleviating the cross-dataset bias between ImageNet and ZSL benchmarks, and ii) reducing the entangled relative geometry relationships between different regions for improving the transferability from seen to unseen classes. These augmented visual features will promote the following sequential learning. To learn locality-augmented visual features, we employ an attribute$\rightarrow$visual decoder in AVT to localize the image regions most relevant to each attribute in a given image (denoted as attribute-based visual features), under the guidance of semantic attribute information. We also take a visual$\rightarrow$attribute decoder to learn visual-based attribute features in VAT. By introducing feature-level and prediction-level semantical collaborative losses further, the two attribute-guided transformers teach each other to further learn two complementary semantic-augmented visual embeddings for key semantic knowledge representations via semantical collaborative learning. Finally, the semantic-augmented visual embeddings cooperated with the semantic vectors are leveraged to conduct desirable visual-semantic interaction for ZSL classification. Extensive experiments show that TransZero++ achieves the new state-of-the-art on three ZSL benchmarks and on the large-scale ImageNet dataset. The qualitative results also demonstrate that TransZero++ refines visual features and accurately localizes attribute regions for semantic-augmented feature representations.
	
	A preliminary version of this work was presented as a conference paper (termed TransZero \cite{Chen2021TransZero}). As shown in Fig. \ref{fig:motivation} (c), although TransZero can localize some important attributes for discriminative region feature representations with low confident scores, some other valuable attributes are failed (\textit{e.g.}, “white wing color” of \textit{Red Legged Kittiwake}).  In this version, we strengthen the work from four aspects: i) We propose VAT to learn visual-based semantic attribute representations that are complementary to the attribute-based visual features learned by AVT, enabling TransZero++ to fully discover the key semantic knowledge from visual features. ii) We introduce feature-level and prediction-level semantic collaborative losses to encourage AVT and VAT to teach each other to discover more intrinsic semantic knowledge for improving the confidence scores of attribute localization, under the guidance of semantic collaborative learning. iii) Since the learned attribute-based visual features and visual-based attribute features are complementary to each other, we fuse the two semantic-augmented visual embeddings learned by AVT and VAT to conduct desirable visual-semantic interaction for ZSL classification. iv) We conduct substantially more experiments to demonstrate the effectiveness of the proposed framework and verify the contribution of each component. Thus, TransZero \cite{Chen2021TransZero} is extended to be TransZero++.

	The main contributions of this paper are summarized as follows:
	\begin{itemize}
		\item  We introduce a novel ZSL method, termed TransZero++, which simultaneously refines the visual features and localizes the object attributes for semantic knowledge representations via semantical collaborative learning. TransZero++ consists of an attribute$\rightarrow$visual Transformer sub-net (AVT) and a visual$\rightarrow$attribute Transformer sub-net (VAT) that learns attribute-based visual features and visual-based attribute features, respectively, which are complementary to each other.

		\item  We propose a feature augmentation encoder to i) alleviate the cross-dataset bias between ImageNet and ZSL benchmarks, and ii) reduce the entangled relative geometry relationships between different regions to improve the transferability of visual features. They are ignored by existing ZSL methods. This feature augmentation encoder is incorporated into AVT and VAT.
		
		\item  We introduce feature-level and prediction-level semantic collaborative losses to enable semantical collaborative learning between the AVT and VAT, encouraging TransZero++ to learn semantic-augmented visual embeddings by discovering more intrinsic semantic knowledge between visual and attribute features.
		
		\item  Extensive experiments demonstrate that TransZero++ achieves the new state-of-the-art on three popular ZSL benchmarks and on the large-scale
		ImageNet dataset. Compared with the popular attention-based method (\textit{i.e.}, APN \cite{Xu2020AttributePN}), TransZero++ leads to significant improvements of 6.3\%/3.2\%, 6.0\%/4.9\% and 4.2\%/8.6\% in $\bm{acc}$/$\bm{H}$ on CUB \cite{Welinder2010CaltechUCSDB2}, SUN \cite{Patterson2012SUNAD} and AWA2 \cite{Xian2019ZeroShotLC}, respectively.
	\end{itemize}
	
	The rest of this paper is organized as follows. Sec. \ref{sec2} discusses related works. The proposed TransZero++ is illustrated in Sec. \ref{sec3}. Experimental results and discussions are provided in Sec. \ref{sec4}. Finally, we present a summary in Sec. \ref{sec5}.

	\begin{figure*}[ht]
		\centering
		\includegraphics[scale=0.315]{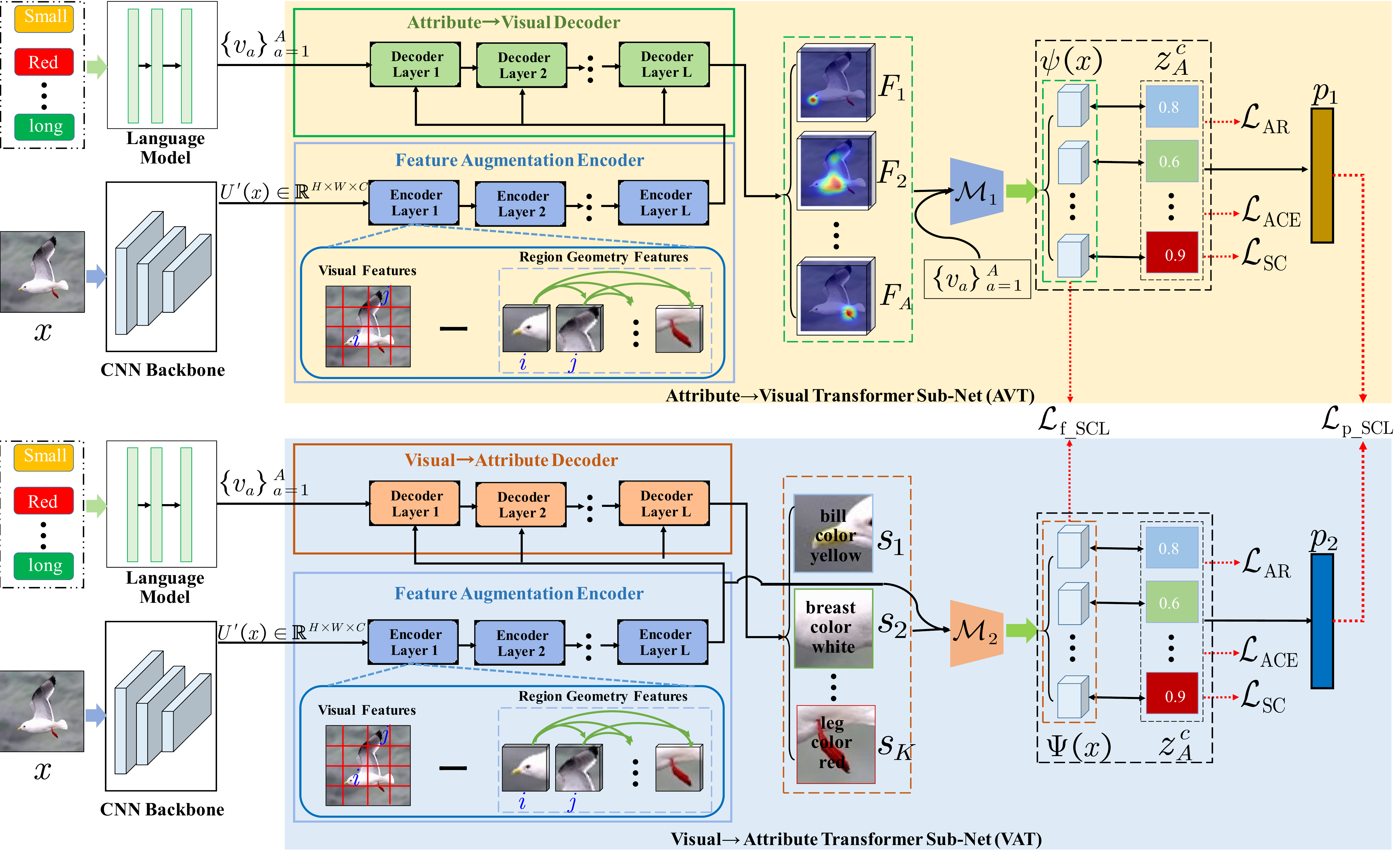}\vspace{-3mm}
		\caption{The architecture of the proposed TransZero++ model. TransZero++ consists of an attribute$\rightarrow$visual Transformer sub-net (AVT) and a visual$\rightarrow$attribute Transformer sub-net (VAT). AVT includes a feature augmentation encoder that alleviates the cross-dataset bias between ImageNet and ZSL benchmarks and reduces the entangled geometry relationships between different regions for improving the transferability of visual features, and an attribute$\rightarrow$visual decoder that localizes object attributes for attribute-based visual feature representations based on the semantic attribute information. Analogously, VAT learns visual-based attribute features using the similar feature augmentation encoder and a visual$\rightarrow$attribute decoder.  Finally, two mapping functions $\mathcal{M}_1$ and $\mathcal{M}_2$ map the learned attribute-based visual features and visual-based attribute features into semantic embedding space respectively under the guidance of semantical collaborative learning, enabling desirable visual-semantic interaction for ZSL classification.}
		\label{fig:pipeline}
	\end{figure*}
	\section{Related Work}\label{sec2}
	In this section, we mainly review three streams of related works: zero-shot learning, Transformer, and collaborative learning.
	\subsection{Zero-Shot Learning}\label{sec2.1} 
	Early embedding-based ZSL methods \cite{Song2018TransductiveUE,Li2018DiscriminativeLO,Han2021ContrastiveEF,Chou2021AdaptiveAG, Han2021ContrastiveEF} aim to learn a mapping from the visual domain to semantic domains to transfer semantic knowledge from seen to unseen classes. They usually extract global visual features from pre-trained or end-to-end trainable networks, \textit{e.g.}, ResetNet \cite{He2016DeepRL}. Note that end-to-end models achieve better performance than pre-trained ones because they fine-tune the visual features, thus the cross-dataset bias between ImageNet and ZSL benchmarks is alleviated \cite{Chen2021FREE,Xian2019FVAEGAND2AF}. 
	
	However, these methods inevitably overfit to seen classes in GZSL since they only learn the model on seen classes \cite{Xian2019ZeroShotLC,Narayan2020LatentEF,Yan2021ZeroNASDG}. As such, the generative ZSL methods are introduced to tackle this challenge using various generative models (\textit{e.g.}, VAEs \cite{Arora2018GeneralizedZL,Schnfeld2019GeneralizedZA,Chen2021HSVA,Chen2021SemanticsDF}, GAN \cite{Arora2018GeneralizedZL,Schnfeld2019GeneralizedZA,Chen2021HSVA}, and generative flows \cite{Shen2020InvertibleZR}) to synthesize a number of images or visual features for unseen classes based on the class semantic vector (attribute values manually annotated by humans). Thus, the ZSL task is converted to supervised classification. Arora \textit{et al.} \cite{Arora2018GeneralizedZL} uses a conditional VAE model (SE-ZSL) to synthesize images for unseen classes. Since synthesizing the high dimensional image is not feasible, Xian \textit{et al.} \cite{Xian2018FeatureGN, Xian2019FVAEGAND2AF} propose f-CLSWGAN and f-VAEGAN to synthesize visual features based on GANs. Different from these generative methods that learn semantic-to-visual mapping as a generator, the common space learning-based ZSL methods are also a special generative model that maps visual and semantic features into a common space simultaneously using VAEs \cite{Tsai2017LearningRV,Schnfeld2019GeneralizedZA, Chen2021SemanticsDF}. 
	
	Although the aforementioned ZSL methods have achieved significant improvements, they still yield relatively undesirable results. This is because they compress holistic visual features to perform global embedding cannot efficiently capture the subtle differences among various fine-grained classes \cite{Huynh2020FineGrainedGZ}. Furthermore, the holistic visual features are limited to poor transferable from one domain to another domain (\textit{e.g.}, from seen to unseen classes) \cite{Atzmon2020ACV,Chen2021CrossDomainFE}.  More relevant to this work are the recent attention-based ZSL methods \cite{Xie2019AttentiveRE,Xie2020RegionGE,Zhu2019SemanticGuidedML,Xu2020AttributePN,Liu2021GoalOrientedGE} that utilize attribute descriptions as guidance to discover the more discriminative region (or part) features. Unfortunately, They simply learn region embeddings (\textit{e.g.}, the whole bird body) neglecting the importance of discriminative attribute localization (\textit{e.g.}, the distinctive bird body parts). Furthermore, the end-to-end attention models are also time-consuming when it comes to fine-tuning the CNN backbone. In contrast, we propose an attribute-guided Transformer to learn the attribute localization for discriminative region feature representations under non end-to-end ZSL model.
	
	\subsection{Transformer Model}\label{sec2.2} 
	Transformer models \cite{Vaswani2017AttentionIA,Naseer2021IntriguingPO,Khan2021TransformersIV,Yu2021MetaFormerIA} have recently achieved excellent performance on a wide range of language and computer vision tasks, \textit{e.g.}, machine translation \cite{Ott2018ScalingNM}, image recognition \cite{Dosovitskiy2021AnII}, video understanding \cite{Gabeur2020MultimodalTF}, visual question answering \cite{ZhangRSTNetCW}, etc. Generally, the success of Transformer can be attributed to its self-supervision and self-attention \cite{Khan2021TransformersIV}. The self-supervision enables complex models to be trained without the high cost of manual annotation, which in turn allows generalizable representations that encode useful relationships between the entities presented in a given dataset to be learned. The self-attention layers consider the broad context of a given sequence by learning the relationships between the elements in the token set  (\textit{e.g.}, words in the language, or patches in an image). Some methods \cite{Gabeur2020MultimodalTF,Cornia2020MeshedMemoryTF,Huang2019AttentionOA,Pan2020XLinearAN} have also shown that the Transformer can better capture the relationship between various modals (\textit{e.g.}, visual features and language) in parallel during training. Motivated by these, we design an attribute-guided Transformer that reduces the relationships among different regions to improve the transferability of visual features and learns the attribute localization for representing discriminative region features. In contrast to most of the vision Transformers that learn feature representations on image patches, our TransZero++ learn semantic-augmented visual embeddings on visual features learned by CNN backbone (\textit{e.g.}, ResNet).
	
	\subsection{Collaborative Learning}\label{sec2.3}
	Recently, Collaborative Learning \cite{Batra2017CooperativeLW} has been introduced to learn multiple models jointly for the same task. Teacher-student models to create consistent training supervisions for labeled/unlabeled data using collaborative learning, enabling two-way knowledge transfer from each other. Thus the intrinsic knowledge between different models is distilled for feature representations \cite{Tarvainen2017MeanTA,Ge2020MutualMP}. Some methods adopt a pool of student models instead of the teacher models by training them with supervision from each other \cite{Zhang2018DeepML,Zhai2020MultipleEB}. These works motivate us to design semantical collaborative learning to discover more intrinsic semantic knowledge (\textit{e.g.}, attributes) for semantic-augmented visual embedding representations on the two attribute-guided Transformers. Different from existing collaborative methods that employ multiple similar networks for implicit knowledge distillation, our semantical collaborative learning is based on two attribute-guided Transformers that learn attribute-based visual features and visual-based attribute features respectively for explicit knowledge distillation.
	
	\section{Proposed Method}\label{sec3}
	First, we introduce some notations and the problem definition of ZSL. We denote $\mathcal{D}^{s}=\left\{\left(x_{i}^{s}, y_{i}^{s}\right)\right\}$ as training data with $C^s$ seen classes, where $x_i^s \in \mathcal{X}$ refers to the image $i$, and $y_i^s \in \mathcal{Y}^s$ is its corresponding class label. The unseen classes $C^u$ have unlabeled samples $\mathcal{D}^{u}=\left\{\left(x_{i}^{u}, y_{i}^{u}\right)\right\}$, where $x_{i}^{u}\in \mathcal{X}$ are the unseen class images, and $y_{i}^{u} \in \mathcal{Y}^u$ are the corresponding labels. A set of class semantic vectors of the class $c \in \mathcal{C}^{s} \cup \mathcal{C}^{u} = \mathcal{C}$ with $A$ attributes $z^{c}=\left[z_{1}^{c}, \ldots, z_{A}^{c}\right]^{\top}= \phi(y)$ (the attribute values are annotated by humans) helps knowledge transfer from seen to unseen classes. According to each word in attribute names, we also take a language model (\textit{i.e.}, GloVe  \cite{Pennington2014GloveGV}) to learn the semantic attribute features $\mathcal{V}_A=\{v_a\}_{a=1}^A$ of all attributes as auxiliary information ($v_a$ is the $a-th$ atribute feature vector). ZSL aims to predict the class labels $y^u \in \mathcal{Y}^u$ in the CZSL settings and $y \in \mathcal{Y}=\mathcal{Y}^{s} \cup \mathcal{Y}^{u}$ in the GZSL setting, where $\mathcal{Y}^{s} \cap \mathcal{Y}^{u}=\emptyset$.
	
	In this paper, we propose a cross attribute-guided Transformer network (termed TransZero++) to refine the visual features, localize the object attributes for discriminative region feature representations, and learn semantic-augmented visual embeddings via semantical collaborative learning under a non end-to-end model. This facilitates desirable visual-semantic interaction in ZSL. As illustrated in Fig. \ref{fig:pipeline}, our TransZero++ includes an attribute$\rightarrow$visual Transformer sub-net (AVT) and visual$\rightarrow$attribute Transformer sub-net (VAT). AVT refines the visual feature using a feature augmentation encoder, and employs an attribute$\rightarrow$visual decoder to learn attribute-based visual features, which is further mapped as the semantic-augmented visual embedding $\psi(x)$ in the semantic embedding space using a mapping function $\mathcal{M}_1$. Analogously, VAT uses a similar feature augmentation encoder and a visual$\rightarrow$attribute decoder to learn visual-based attribute features, which are further mapped as the semantic-augmented visual embeddings $\Psi(x)$ in the semantic embedding space using another mapping function $\mathcal{M}_2$. Finally, the two semantic-augmented visual embeddings are fused to conduct desirable visual-semantic interaction for ZSL classification based on the class semantic vectors. To encourage TransZero++ to learn semantic-augmented visual embeddings, we introduce feature-level and prediction-level semantical collaborative losses to encourage the two cross AVT and VAT to learn collaboratively and teach each other throughout the training process. 
	
	\begin{figure*}[t]
		\begin{center}
			\includegraphics[scale=0.35]{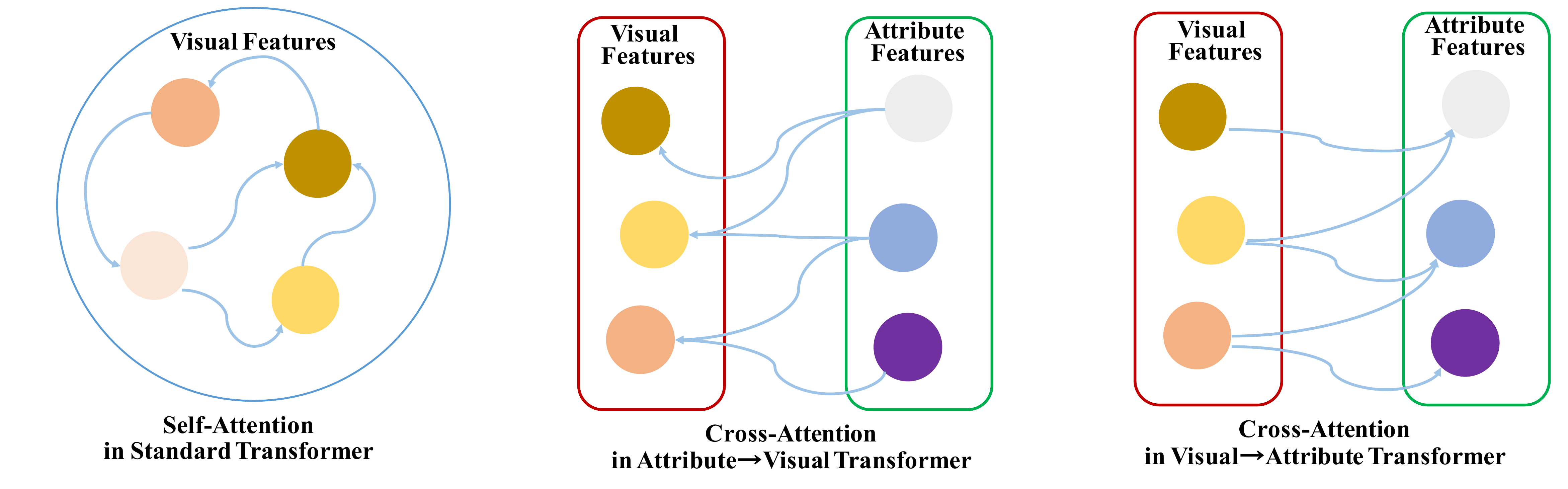}\vspace{-3mm}
			\caption{The self-attention operator in standard Transformer, and the cross-attention operator in our  attribute$\rightarrow$visual Transformer and visual$\rightarrow$attribute Transformer.}
			\label{fig:atten-operator}
		\end{center}
	\end{figure*}

	\subsection{Attribute$\rightarrow$Visual Transformer}\label{sec3.1}
	\subsubsection{Feature Augmentation Encoder.}\label{sec3.1.1}
	Since the cross-dataset bias between ImageNet and ZSL benchmarks potentially limits the quality of visual feature extraction \cite{Chen2021FREE,Chen2021SemanticsDF}, we first propose a feature augmentation encoder (FAE) to refine the visual features of ZSL benchmarks. In addition, previous ZSL methods simply flatten the grid features $U^{\prime}(x)\in \mathbb{R}^{H \times W \times C}$ (extracted by a CNN backbone) of a single image into a feature vector, which is further applied to generative models or embedding learning. Unfortunately, such a feature vector implicitly entangles the visual feature representations among various regions in an image, which hinders their transferability from one domain to other domains (e.g., from seen to unseen classes) \cite{Xu2020AttributePN,ZhangRSTNetCW}. Atzmon \textit{et al.} \cite{Atzmon2020ACV} and Chen \cite{Chen2021CrossDomainFE} show that the local visual features are more transferable than the holistic ones. As such, we propose a feature-augmented scaled dot-product attention to further enhance the encoder layer by reducing the relative geometry relationships among the grid features.
	
	Motivated by \cite{Herdade2019ImageCT}, we first calculate the relative center coordinates ($v^{\text{cen}}_i$, $t^{\text{cen}}_i$) based on the pair of 2-D relative positions of the $i$-th grid $\left\{\left(v_{i}^{\min }, t_{i}^{\min }\right),\left(v_{i}^{\max }, t_{i}^{\max }\right)\right\}$ during learning the relative geometry features, formulated as:
	\begin{align}
	\label{eq:center-coor}
	&\left(v^{\text{cen}}_{i}, t^{\text{cen}}_{i}\right)=\left(\frac{v_{i}^{\min }+v_{i}^{\max }}{2}, \frac{t_{i}^{\min }+t_{i}^{\max }}{2}\right),\\
	&w_{i}=\left(v_{i}^{\max }-v_{i}^{\min }\right)+1,\\
	&h_{i}=\left(t_{i}^{\max }-t_{i}^{\min }\right)+1,
	\end{align}
	where $(v_i^{\min }, t_i^{\min })$ and $(v_i^{\max }, t_i^{\max })$ are the relative position coordinates of the top left corner and bottom right corner of the grid $i$, respectively. Different from \cite{Herdade2019ImageCT} that uses 4-D feature vectors for geometry feature representations, we only need to calculate the 2-D geometry feature representations as our grid features are irrelevant to the edges (\textit{i.e.}, width and length) of grid. This is because \cite{Herdade2019ImageCT} learn the bounding boxes with various shapes while our grid visual features are shared with same shape.
	
	Different to \cite{Herdade2019ImageCT}, it attempts to take advantage of the geometric relationship representations for improving the caption performance, by incorporating relative geometry features into the attention weight matrix. In contrast, our FAE aims to remove the relative geometry prior of various image regions from the visual feature representations. Thus, the transferability and discrimination of visual features are enhanced. Specifically, we construct region geometry features $G_{i j}$ between grid $i$ and grid $j$:
	\begin{gather}
	\label{eq:geometry-repre}
	G_{i j}=\operatorname{ReLU}\left(w_{g}^{T} g_{i j}\right),
	\end{gather}
	where
	\begin{gather}
	g_{i j}=F C\left(r_{i j}\right), \qquad r_{i j}=\left(\begin{array}{c}
	\log \left(\frac{\left|v^{\text{cen}}_{i}-v^{\text{cen}}_{j}\right|}{w_{i}}\right) \\
	\log \left(\frac{\left|t^{\text{cen}}_{i}-t^{\text{cen}}_{j}\right|}{h_{i}}\right) \\
	\end{array}\right),
	\end{gather}
	where $r_{ij}$ is the relative geometry relationship between grids $i$ and $j$, $FC$ is a fully connected layer followed by a $ReLU$ activation, and $w_{g}^{T}$ is a set of learnable weight parameters.
	
	Finally, we substract the region geometry features from the visual features in the feature-augmented scaled dot-product attention to provide a more accurate attention map, formally defined as: 
	\begin{align}
	\label{eq:encoder-atten}
	&Q^e=U(x) W_{q}^e, K^e=U(x) W_{k}^e, V^e=U(x) W_{v}^e, \\
	&Z_{a u g}=\operatorname{softmax}\left(\frac{Q^e K^{e^{\top}}}{\sqrt{d^e}}-G\right) V^e,\\
	&U_{aug}(x) \leftarrow U(x)+Z_{aug},
	\end{align}
	where $Q$, $K$, $V$ are the query, key, and value matrices, $W_q^e$, $W_k^e$, $W_v^e$ are the learnable matrices of weights, $d^e$ is a scaling factor, and $Z_{aug}$ is the augmented features. $U(x)\in \mathbb{R}^{C \times HW}$ are the packed visual features, which are learned from the flattened features embedded by a fully connected layer followed by a ReLU and a Dropout layer. $U_{aug}(x)$ is the augmented visual features from the feature augmentation encoder, they will promote the following sequential learning.We rewrite $U_{aug}(x)$ as $U_{aug}^{a \rightarrow v} (x)$ and $U_{aug}^{v \rightarrow a}(x)$ in AVT and VAT, respectively.

	\subsubsection{Attribute$\rightarrow$Visual Decoder.}\label{sec3.1.2}
	To learn attribute-based visual features, we design attribute$\rightarrow$visual decoder to localize the most relevant image regions to the corresponding attributes to extract attribute-based visual features from a given image for each attribute. We can attend to image regions with respect to each attribute, and compare each attribute to the corresponding attended visual region features to determine the importance of each attribute. In the standard Transformer \cite{Vaswani2017AttentionIA}, it takes the \textit{self-attention} operator to consider all pairwise relations among the visual feature elements. The operator adjusts every single element by attending them to the others. (as shown in Fig. \ref{fig:atten-operator} (Left)). However, our \textit{cross-attention} operator attends to visual features from attribute features, as shown in Fig. \ref{fig:atten-operator} (Middle). In the decoding process, the attribute$\rightarrow$visual decoder continuously localizes the local visual information under the guidance of semantic attribute features $\mathcal{V}_A$. Thus, our attribute$\rightarrow$visual decoder can effectively localize the image regions most relevant to each attribute in a given image. The multi-head cross-attention layer uses the outputs of the encoder $U_{aug}^{a\rightarrow v}(x)$ as keys ($K_t^{a\rightarrow v}$) and values ($V_t^{a\rightarrow v}$) and a set of learnable semantic embeddings $\mathcal{V}_A$ as queries ($Q_t^{a\rightarrow v}$). It is defined as:
	\begin{align}
	\label{eq:decoder-attri}
	&Q_t^{a\rightarrow v}=\mathcal{V}_A W_{qt}^{a\rightarrow v},\\
	&K_t^{a\rightarrow v}=U_{aug}^{a\rightarrow v}(x) W_{kt}^{a\rightarrow v},\\
	&V_t^{a\rightarrow v}=U_{aug}^{a\rightarrow v}(x) W_{vt}^{a\rightarrow v}, \\
	&\text{head}_{t}=\operatorname{softmax}\left(\frac{Q_t^d K_t^{{a\rightarrow v}^{\top}}}{\sqrt{\tau}}\right) V_t^{a\rightarrow v},\\
	&\hat{F}= \|_{t=1}^{T} (\text{head}_{t})W_o^{a\rightarrow v},
	\end{align}
	where $W_{qt}^{a\rightarrow v}$, $W_{kt}^{a\rightarrow v}$, $W_{vt}^{a\rightarrow v}$, and $W_o^{a\rightarrow v}$ are the learnable weights, $\tau$ is a scaling factor, and $\|$ is a concatenation operation. Thus, we get a set of attribute-based visual features $\hat{F}=\{\hat{F}_1,\cdots,\hat{F}_A\}$, which captures the visual evidence used to localize the corresponding semantic attributes in the image. Specifically, our AVT will assign a high positive score to the $a$-th attribute if an image has an obvious attribute $v_a$. Otherwise, AVT will assign a low score to the $a$-th attribute. Then, a feed-forward network (FFN) with two linear transformations followed a ReLU activation in between is applied to the attended features $\hat{F}$:
	\begin{gather}
	\label{eq:FFN-av}
	F=ReLu \left(\hat{F} W_{1}^{a\rightarrow v}+b_{1}^{a\rightarrow v}\right) W_{2}^{a\rightarrow v}+b_{2}^{a\rightarrow v},
	\end{gather}
	where $W_{1}^{a\rightarrow v}$, $W_{2}^{a\rightarrow v}$, $b_1^{a\rightarrow v}$ and $b_2^{a\rightarrow v}$ are the weights and biases of the linear layers respectively, and $F=\{F_1,\cdots,F_A\}$ are the final attribute-based visual features that will be mapped into semantic embedding space for desirable visual-semantic interaction using a mapping function ($\mathcal{M}_1$).

	\subsubsection{Visual-Semantic Embedding Mapping}\label{sec3.1.3}
	After learning attribute-based visual features that are locality-augmented, we further map them into the semantic embedding space. Based on a mapping function ($\mathcal{M}_1$), we take the semantic attribute vectors $\mathcal{V}_A=\{v_a\}_{a=1}^A$ as support to encourage the mapping to be more accurate. Specifically, $\mathcal{M}_1$ matches the attribute-based visual features $F$ with the semantic attribute information $v_{a}$, formulated as:
	\begin{gather}
	\label{Eq:av_encoder}
	\psi(x_i)=\mathcal{M}_1(F)= \mathcal{V}_A^{\top} W_3^{a\rightarrow v} F,
	\end{gather}
	where $W_3^{a\rightarrow v}$ is an embedding matrix that embeds $F$ into the semantic attribute space. Similar to the class semantic vector $z^c$, $\psi_a(x_i)$ is an attribute score that represents the confidence of having the $a$-th attribute in the image $x_i$. Given a set of semantic attribute vectors $\mathcal{V}_A=\{v_{a}\}_{a=1}^A$, TransZero++ obtains a mapped semantic-augmented visual embedding $\psi(x_i)$ of a single image $x_i$.
	
	\subsection{Visual$\rightarrow$Attribute Transformer}\label{sec3.2}
	Likewise, we introduce visual$\rightarrow$attribute Transformer (VAT) to attend to attributes with respect to each image region, and thus the visual-based attribute features are learned. They are complementary to the attribute-based visual features, enabling them to calibrate each other to discover more intrinsic semantic knowledge between visual and attribute features. VAT first applies the similar feature augmentation encoder to refine the visual features as $U_{aug}^{v\rightarrow a}(x)$, which are further used in visual$\rightarrow$attribute decoder of VAT. 
	
	\subsubsection{Visual$\rightarrow$Attribute Decoder.}\label{sec3.2.1}
	After refining the visual features, we design a visual$\rightarrow$attribute decoder {\color{blue}to learn} visual-based attribute features. Specifically, we take the cross-attention operator to attend the attribute from visual representations, as shown in Fig. \ref{fig:atten-operator} (Right). Formally, it is formulated as:
	\begin{align}
	\label{eq:decoder-vis}
	&Q_t^{v\rightarrow a}= U_{aug}^{v\rightarrow a}(x) W_{qt}^{v\rightarrow a},\\
	&K_t^{v\rightarrow a}=\mathcal{V}_A W_{kt}^{v\rightarrow a}, \\
	&V_t^{v\rightarrow a}=\mathcal{V}_A W_{vt}^{v\rightarrow a}, \\
	&\text{head}_{t}=\operatorname{softmax}\left(\frac{Q_t^d K_t^{{v\rightarrow a}^{\top}}}{\sqrt{\tau}}\right) V_t^{v\rightarrow a},\\
	&\hat{S}= \|_{t=1}^{T} (\text{head}_{t})W_o^{v\rightarrow a},
	\end{align}
	where $W_{qt}^{v\rightarrow a}$, $W_{kt}^{v\rightarrow a}$, $W_{vt}^{v\rightarrow a}$, and $W_o^{v\rightarrow a}$ are the learnable weights, and $\|$ is a concatenation operation. As such, we get a set of visual-based attribute features $\hat{S}=\{\hat{S}_1,\cdots,\hat{S}_K\}$. Intrinsically, $\hat{S}$ is the visual semantic representations corresponding to the $K=H\times W$ visual regions in a single image. Specifically, our VAT will assign a high positive score to the $k$-th visual region with respect to the corresponding attribute, otherwise, VAT will assign a low score. Similar to the AVT, an FFN is utilized to the attended features $\hat{S}$:
	\begin{gather}
	\label{eq:FFN-va}
	S=ReLu \left(\hat{S} W_{1}^{v\rightarrow a}+b_{1}^{v\rightarrow a}\right) W_{2}^{v\rightarrow a}+b_{2}^{v\rightarrow a},
	\end{gather}
	where $W_{1}^{v\rightarrow a}$, $W_{2}^{v\rightarrow a}$, $b_1^{v\rightarrow a}$ and $b_2^{v\rightarrow a}$ are the weights and biases of the linear layers respectively, and $S=\{S_1,\cdots,S_K\}$ are the final visual-based attribute features, which will be mapped into semantic embedding space for significant visual-semantic interaction using a mapping function $\mathcal{M}_2$.

	\subsubsection{Visual-Semantic Embedding Mapping}\label{sec3.2.2}
	Once visual-based attribute features are learned, we map them into the semantic embedding space based on a mapping function ($\mathcal{M}_2$). To conduct an effective map, we take the augmented visual features $U_{aug}^{v\rightarrow a}(x)$ learned by feature augmentation encoder as support. Thus, $\mathcal{M}_2$ first maps the visual-based attribute features $S$ into $K$ region scores $\bar{S}$, formulated as:
	\begin{gather}
	\label{Eq:va_encoder}
	\bar{S}=\mathcal{M}_2(S)= U_{aug}^{v\rightarrow a}(x)^{\top} W_3^{v\rightarrow a} S,
	\end{gather}
	where $W_3^{v\rightarrow a}$ is a learnable mapping matrix. Here, $\bar{S}$ is $K$-D, which is not matched with the dimension of class semantic vector $A$-D. Thus, $\mathcal{M}_2$ further embeds $\bar{S}$ into the semantic attribute space with dimension of $A$ based on an attention score $Att=\mathcal{V}_A^{\top} W_{att}U(x) \in \mathbb{R}^{A \times K}$, where $W_{att}$ is a learnable matrix, written as:
		\begin{gather}
		\label{Eq:va_enbed}
		\Psi(x_i)= Att \times \bar{S} ,
		\end{gather}
	Similar to the $\psi(x_i)$, $\Psi_a(x_i)$ is an attribute score that represents the confidence of having the $a$-th attribute in the image $x_i$. As such, TransZero++ obtains a mapped semantic-augmented visual embedding $\Psi(x_i)$ of a single image $x_i$ in VAT.

	\subsection{Model Optimization}\label{sec3.4}
	To achieve effective optimization for TransZero++, each attribute-guided Transformer sub-net is trained with three supervised losses that have been used in our conference version \cite{Chen2021TransZero}, \textit{i.e.}, the attribute regression loss, attribute-based cross-entropy loss, and self-calibration loss. To enable semantic collaborative learning between the two attribute-guided Transformer sub-nets, \textit{i.e.}, AVT and VAT,  we introduce a feature-level semantic collaborative loss and prediction-level semantic collaborative loss, which align each other's visual embeddings and class posterior probabilities respectively.
	
	\noindent\textbf{Attribute Regression Loss.}
	To encourage $\mathcal{M}_1$ and $\mathcal{M}_2$ to accurately map visual/attribute features into their corresponding class semantic vectors, we introduce an attribute regression loss to constrain TransZero++. Here, we regard visual-semantic mapping as a regression problem and minimize the mean square error between the embedded attribute score $f(x_i)$ and the corresponding ground truth attribute score $z^{c}$  of a batch of $n_b$ images $\{x_i^s\}_{i=1}^{n_b}$:
	\begin{gather}
	\label{eq:reg-loss}
	\mathcal{L}_{\text{AR}}=\frac{1}{n_{b}} \sum_{i=1}^{n_{b}}\|f(x_i^s)-z^{c}\|_{2}^{2}.
	\end{gather}
	where $f(x_i^s)=\psi(x_i^s)$ for AVT and $f(x_i^s)=\Psi(x_i^s)$ for VAT.
	
	\noindent\textbf{Attribute-Based Cross-Entropy Loss.}
	Since the associated visual/attribute embedding is projected near its class semantic vector $z^{c}$ when an attribute is visually present in an image, we take the attribute-based cross-entropy loss $\mathcal{L}_{\text{ACE}}$ to optimize the parameters of the TransZero++, i.e., the dot product between the visual/attribute embedding and each class semantic vector is calculated to produce class logits. This encourages the image/attribute to have the highest compatibility score with its corresponding class semantic vector. Given a batch of $n_b$ training images $\{x_i^s\}_{i=1}^{n_b}$ with their corresponding class semantic vectors $z^c$,  $\mathcal{L}_{\text{ACE}}$ is defined as:
	\begin{gather}
	\label{eq:L_ACE}
	\mathcal{L}_{\text{ACE}}=-\frac{1}{n_{b}} \sum_{i=1}^{n_{b}} \log \frac{\exp \left(f(x_i^s) \times z^{c}\right)}{\sum_{\hat{c} \in \mathcal{C}} \exp \left(f(x_i^s) \times z^{\hat{c}} \right)}.
	\end{gather}

	\begin{figure*}[ht]
		\begin{center}
			\includegraphics[width=1\linewidth]{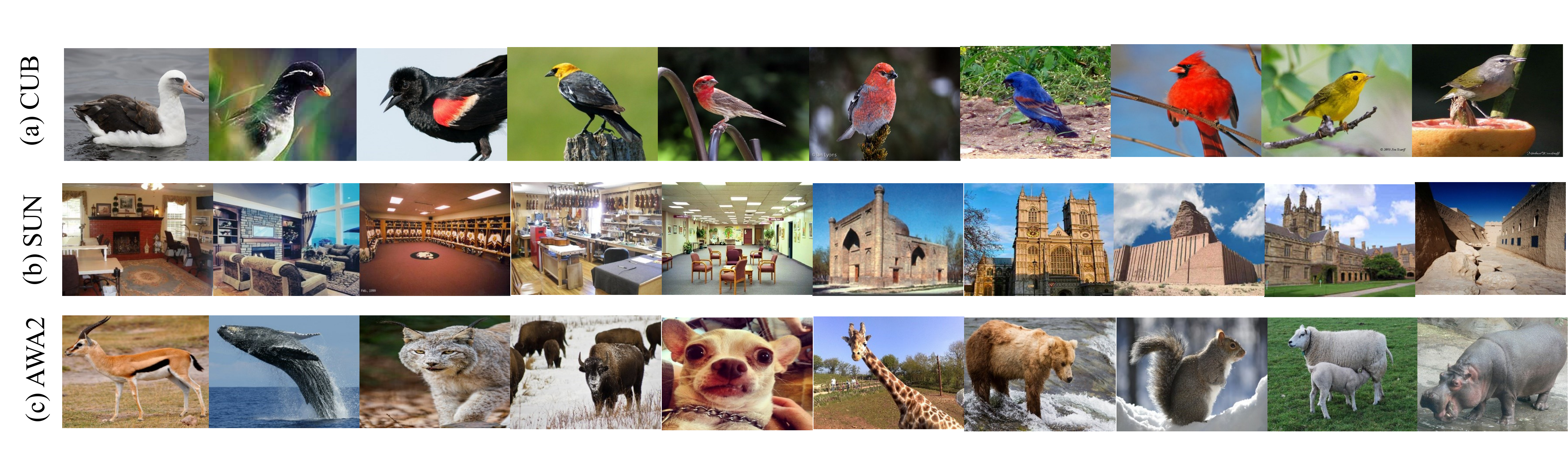}\vspace{-5mm}
			\caption{{Some samples on various datasets}, including two fine-grained datasets (\textit{i.e.}, CUB and SUN), and one coarse-grained dataset (\textit{i.e.}, AWA2). Each sample is extract from various classes. (Best viewed in color.)}
			\label{fig:samples}
		\end{center}
	\end{figure*}
	
	\noindent\textbf{Self-Calibration Loss.}
	Since $\mathcal{L}_{\text{AR}}$ and $\mathcal{L}_{\text{ACE}}$ optimize the model only on seen classes, TransZero++ inevitably overfits to these classes  \cite{Zhu2019SemanticGuidedML,Huynh2020FineGrainedGZ,Xu2020AttributePN}. To tackle this challenge, we further employ a self-calibration loss $\mathcal{L}_{\text{SC}}$ to explicitly shift some of the prediction probabilities from seen to unseen classes. $\mathcal{L}_{\text{SC}}$ is thus formulated as:
	\begin{gather}
	\label{eq:L_SC}
	\mathcal{L}_{\text{SC}}=-\frac{1}{n_{b}} \sum_{i=1}^{n_{b}}   \sum_{c^{\prime=1}}^{\mathcal{C}^u} \log \frac{\exp \left(f(x_i^s) \times z^{c^{\prime}} + \mathbb {I}_{\left[c^{\prime}\in\mathcal{C}^u\right]}\right)}{\sum_{\hat{c} \in \mathcal{C}} \exp \left(f(x_i^s) \times z^{\hat{c}} + \mathbb {I}_{\left[\hat{c}\in\mathcal{C}^u\right]}\right)},
	\end{gather}
	where $\mathbb {I}_{\left[c\in \mathcal{C}^u\right]}$ is an indicator function (\textit{i.e.}, it is 1 when $c\in\mathcal{C}^u$, otherwise -1). Intuitively, $\mathcal{L}_{\text{SC}}$ encourages non-zero probabilities to be assigned to the unseen classes during training, which allows TransZero++ to produce a large/non-zero probability for the true unseen class when given test samples from unseen classes.

	\noindent\textbf{Semantical Collaborative Loss.}
	To enable the two attribute-augmented Transformer sub-nets to learn collaboratively and teach each other throughout the training process, we further introduce a feature-level semantical collaborative loss $\mathcal{L}_{\text{f\_SCL}}$ and a prediction-level semantical collaborative loss $\mathcal{L}_{\text{p\_SCL}}$ for optimization. These two losses are based on $\ell_2$ distance. Note that the $\ell_2$ distance can be replaced with other metrics, \textit{e.g.}, the Kullback Leibler (KL) Divergence or Jensen-Shannon Divergence (JSD).

	Specifically, $\mathcal{L}_{\text{f\_SCL}}$ uses an $\ell_2$ distance between the semantic-augmented visual embeddings of AVT and VAT (\textit{i.e.}, $\psi(x_i)$ and $\Psi(x_i)$) for test sample $x_i$, formulated as:
	\begin{gather}
	\label{eq:f_SCL}
	\mathcal{L}_{\text{f\_SCL}}=\frac{1}{n_{b}} \sum_{i=1}^{n_{b}}\|\psi(x_i)-\Psi(x_i)\|_{2}^{2}.
	\end{gather}
	
	Likewise, $\mathcal{L}_{\text{p\_SCL}}$ calculates the $\ell_2$ distance between the predictions of the two attribute-augmented Transformer sub-nets (\textit{i.e.}, $p_1$ and $p_2$), formulated as:
	\begin{gather}
	\label{eq:p_SCL}
	\mathcal{L}_{\text{p\_SCL}}=\frac{1}{n_{b}} \sum_{i=1}^{n_{b}}\|p_1(x_i)-p_2(x_i)\|_{2}^{2},
	\end{gather}
	
	\begin{algorithm}[t]
		\caption{The algorithm of TransZero++.}
		\label{algotithm:TransZero}
		\begin{algorithmic}[1]
			\Require  The training set $\mathcal{D}^{s}=\left\{\left(x_{i}^{s}, y_{i}^{s}\right)\right\}_{i=1}^{N^s}$, the test set $\mathcal{D}^{u}=\left\{\left(x_{i}^{u}, y_{i}^{u}\right)\right\}_{i=1}^{N^u}$, the pretrained CNN backbone ResNet101, the maximum iteration epoch $\mathrm{max}_{iter}$, loss weights (\textit{i.e.}, $\mathcal{L}_{\text{AR}}$, $\mathcal{L}_{\text{SC}}$, $\lambda_{\text{VAT}}$, $\lambda_{\text{f\_SCL}}$ and $\lambda_{\text{p\_SCL}}$), the combination coefficients $\alpha$, and hyperparameters (learning rate = 0.001, betas = (0.5, 0.999)) of the Adam optimizer.
			\Ensure The predicted label $c^*$ for the test samples.
			\While{$iter \leq  \mathrm{max}_{iter}$} 	\Comment{\textit{\color{gray} Optimization}}
			\State Take CNN backbone (\textit{e.g.}, ResNet101 \cite{He2016DeepRL}) to extract the visual features $U(x)$ for all image samples.
			\State Take language model (\textit{i.e.}, GloVe  \cite{Pennington2014GloveGV}) to learn the semantic attribute features $\mathcal{V}_A=\{v_1,\cdots,v_a\}_{a=1}^{A}$ for each attribute.
			\State Optimize TransZero++ with Eq. \ref{Eq:L_final}.
			\EndWhile
			\State Predict the label $c^*$ of the test samples using Eq. \ref{eq:prediction}. \Comment{\textit{\color{gray} Prediction}}
		\end{algorithmic}
	\end{algorithm}

	\begin{table*}[ht]
		\centering  
		\caption{Results ~($\%$) of the state-of-the-art CZSL and GZSL modes on CUB, SUN and AWA2, including end-to-end and non end-to-end methods (generative and non-generative methods). The best and second-best results are marked in \textbf{bold} and \underline{underline}, respectively. The Symbol “--” indicates no results. The symbol “*” denotes attention-based methods. The symbol “$\dagger$” denotes the methods using calibration. The symbol “$\ddagger$” denotes the methods using finetuned features.}\label{Table:SOTA}\vspace{-2mm}
		\resizebox{1.0\linewidth}{!}{
			\begin{tabular}{r|c|ccc|c|ccc|c|ccc}
				\hline
				\multirow{3}{*}{\textbf{Methods}} 
				&\multicolumn{4}{c|}{\textbf{CUB}}&\multicolumn{4}{c|}{\textbf{SUN}}&\multicolumn{4}{c}{\textbf{AWA2}}\\
				\cline{2-5}\cline{6-9}\cline{9-13}
				&\multicolumn{1}{c|}{CZSL}&\multicolumn{3}{c|}{GZSL}&\multicolumn{1}{c|}{CZSL}&\multicolumn{3}{c|}{GZSL}&\multicolumn{1}{c|}{CZSL}&\multicolumn{3}{c}{GZSL}\\
				\cline{2-5}\cline{6-9}\cline{9-13}
				\textbf{} 
				&\rm{acc}&\rm{U} & \rm{S} & \rm{H} &\rm{acc}&\rm{U} & \rm{S} & \rm{H} &\rm{acc}&\rm{U} & \rm{S} & \rm{H} \\
				\hline 
				\textbf{End-to-End} &&&&&&&&&&&&\\   
				\rowcolor{mygray}QFSL~\cite{Song2018TransductiveUE}&58.8&33.3&48.1&39.4&56.2&30.9&18.5&23.1&63.5&52.1&72.8&60.7\\
				LDF~\cite{Li2018DiscriminativeLO} &67.5&26.4&\textbf{81.6}&39.9&--&--&--&--&65.5&9.8& 87.4& 17.6 \\
				\rowcolor{mygray}SGMA$^{*}$~\cite{Zhu2019SemanticGuidedML} &71.0&36.7&71.3&48.5&--&--&--&--&68.8&37.6&87.1&52.5\\
				AREN$^{*}$~\cite{Xie2019AttentiveRE}&71.8&38.9&78.7&52.1&60.6&19.0&38.8&25.5&67.9&15.6&\underline{92.9}&26.7 \\
				\rowcolor{mygray}LFGAA$^{*}$~\cite{Liu2019AttributeAF}&67.6&36.2&\underline{80.9}&50.0&61.5&18.5&40.0&25.3&68.1&27.0&\textbf{93.4}&41.9\\
				APN$^{*\dagger}$~\cite{Xu2020AttributePN}&72.0&65.3& 69.3&67.2&61.6& 41.9&34.0&37.6&68.4&57.1&72.4&63.9\\
				\rowcolor{mygray}{GEM-ZSL$^{*\dagger}$}~\cite{Liu2021GoalOrientedGE}&\underline{77.8}& 64.8& 77.1& \textbf{70.4}& 62.8& 38.1& 35.7& 36.9& 67.3& \textbf{64.8}& 77.5& 70.6\\
				
				\hline
				\textbf{Non End-to-End} &&&&&&&&&&&&\\ 
				\textbf{\textit{Generative Methods}} &&&&&&&&&&&&\\ 
				\rowcolor{mygray}SE-ZSL~\cite{Arora2018GeneralizedZL} &59.6& 41.5 & 53.3 & 46.7& 63.4 & 40.9 & 30.5 & 34.9 & 69.2 & 58.3 & 68.1 & 62.8 \\
				f-CLSWGAN~\cite{Xian2018FeatureGN}    &57.3&43.7&57.7& 49.7&60.8&42.6&36.6&39.4&68.2&57.9&61.4&59.6\\
				\rowcolor{mygray}f-VAEGAN~\cite{Xian2019FVAEGAND2AF}&61.0&48.4&60.1& 53.6&64.7&45.1&38.0&41.3&71.1&57.6&70.6&63.5\\
				{OCD-CVAE~\cite{Keshari2020GeneralizedZL}}&--&44.8&59.9&51.3&--&44.8&42.9&43.8&--&59.5&73.4&65.7\\
				\rowcolor{mygray}Composer~\cite{Huynh2020CompositionalZL}&69.4&56.4&63.8&59.9&62.6& \textbf{55.1}&22.0& 31.4&71.5&62.1&77.3&68.8\\
				TF-VAEGAN~\cite{Narayan2020LatentEF}& 64.9&52.8&64.7&58.1&66.0&45.6&40.7& 43.0& 72.2& 59.8& 75.1& 66.6 \\
				\rowcolor{mygray}{IZF~\cite{Shen2020InvertibleZR}} & 67.1 & 52.7 & 68.0 & 59.4 & \textbf{68.4} & \underline{52.7} & \underline{57.0} & \underline{54.8} & \textbf{74.5} & 60.6 & 77.5 & 68.0 \\
				{GCM-CF~\cite{Yue2021CounterfactualZA}}&--&61.0&59.7&60.3&--& 47.9&37.8& 42.2&--& 60.4&75.1&67.0\\
				\rowcolor{mygray}SDGZSL~\cite{Chen2021SemanticsDF}&75.5&59.9&66.4&63.0&--& --&--& --&72.1& \underline{64.6}&73.6&68.8\\
				FREE~\cite{Chen2021FREE}&--&55.7&59.9&57.7&--& 47.4&37.2& 41.7&--& 60.4&75.4&67.1\\
				\rowcolor{mygray}HSVA~\cite{Chen2021HSVA}&62.8&52.7&58.3&55.3&63.8&48.6&39.0&43.3&--&59.3&76.6&66.8\\
				LBP~\cite{Lu2021ZeroAF}& 61.9& 42.7& 71.6 & 53.5 & 63.2 & 39.2 & 36.9 & 38.1 &--&--&--&--\\
				\rowcolor{mygray}{FREE+ESZSL~\cite{Cetin2022CL}}&--& 51.6& 60.4& 55.7&--&48.2&36.5&41.5&--&51.3&78.0& 61.8\\
				{APN+f-VAEGAN-D2} $^\ddagger$~\cite{Xu2022AttributePN}&73.9&65.5&75.6 &\underline{70.2}&65.9&41.4 &\textbf{89.9}& \textbf{56.7}&71.2&63.2 &81.0& 71.0\\
				
				\cdashline{2-13}[4pt/1pt]
				\textbf{\textit{Non-Generative Methods}} &&&&&&&&&&&&\\ 
				\rowcolor{mygray}SP-AEN~~\cite{Chen2018ZeroShotVR}      &55.4&34.7&70.6&46.6 &59.2&24.9&38.6&30.3&58.5&23.3&90.9&37.1 \\
				PQZSL~\cite{Li2019CompressingUI}     &--&43.2&51.4&46.9 &--&35.1& 35.3&35.2&--&31.7& 70.9& 43.8 \\
				\rowcolor{mygray}IIR$^{\dagger}$~\cite{Cacheux2019ModelingIA}&63.8&55.8&52.3& 53.0&63.5&47.9&30.4& 36.8&67.9&48.5&83.2& 61.3\\
				TCN~\cite{Jiang2019TransferableCN}   &59.5&52.6&52.0&52.3&61.5&31.2&37.3&34.0&71.2&61.2&65.8&63.4\\
				\rowcolor{mygray}DVBE~\cite{Min2020DomainAwareVB}&--&53.2&60.2&56.5&--&45.0&37.2&40.7&--&63.6&70.8&67.0\\ 
				DAZLE$^{*\dagger}$~\cite{Huynh2020FineGrainedGZ}&66.0&56.7&59.6&58.1&59.4&52.3&24.3&33.2&67.9&60.3&75.7&67.1\\
				GNDAN$^{*\dagger}$~\cite{Chen2022GNDANGN}&75.1&\underline{69.2}&69.6&69.4&65.3&50.0&34.7&41.0&71.0&60.2&80.8&69.0\\
				MSDN$^{*\dagger}$~\cite{Chen2022MSDNMS}&76.1&68.7&67.5&68.1&65.8&52.2&34.2&41.3&70.1&62.0&74.5&67.7\\
				\rowcolor{mygray}TransZero~\cite{Chen2021TransZero}(Conference Version) &76.8&\textbf{69.3}&68.3&68.8&65.6&52.6&33.4&40.8&70.1&61.3&82.3&\underline{70.2}\\
				\hline
				{\textbf{TransZero++}}{~\textbf{(Ours)}}    &\textbf{78.3}&67.5&73.6&\textbf{70.4}&\underline{67.6}&48.6&37.8&42.5&\underline{72.6}&\underline{64.6}&82.7&\textbf{72.5}\\
				\hline	
		\end{tabular} }
		\label{table:sota}
	\end{table*}
	
	\begin{table*}[ht]
		\centering
		\caption{Conventional zero-shot learning results of various methods on ImageNet. Results indicated with $*$ and $\dagger$ are taken from \cite{Kampffmeyer2019RethinkingKG} and \cite{Hascoet2019OnZR}, respectively.} \label{table:ImageNet}
		\vspace{-2mm}
		\resizebox{1.0\linewidth}{!}{
			{
				\begin{tabular}{l|ccccccc}
					\hline
					Methods&ConSE$^*$\cite{Norouzi2014ZeroShotLB}&SYNC$^*$\cite{Changpinyo2016SynthesizedCF} &EXEM$^*$\cite{Changpinyo2017PredictingVE}& GCNZ$^*$\cite{Wang2018ZeroShotRV} & Trivial$^\dagger$\cite{Hascoet2019OnZR}&SGCN$^*$\cite{Kampffmeyer2019RethinkingKG}&TransZero++\\
					\hline
					Top-1 (\%) & 8.30&10.5& 12.5&19.8&20.3&26.2&23.9\\
					\hline
			\end{tabular}}
		}
	\end{table*}
	
	Similar to the TransZero, the AVT and VAT are optimized with the three supervised losses, \textit{i.e.}, $\mathcal{L}_{\text{ACE}}$,  $\mathcal{L}_{\text{AR}}$ and $\mathcal{L}_{\text{SC}}$, formulated as:
	\begin{align}
	\label{Eq:L_AVT}
	&\mathcal{L}_{\text{AVT}} =  \mathcal{L}_{\text{ACE}}^{\text{AVT}} + \lambda_{\text{AR}}\mathcal{L}_{\text{AR}}^{\text{AVT}}+ \lambda_{\text{SC}}\mathcal{L}_{\text{SC}}^{\text{AVT}},\\
	&\mathcal{L}_{\text{VAT}}=  \mathcal{L}_{\text{ACE}}^{\text{VAT}} + \lambda_{\text{AR}}\mathcal{L}_{\text{AR}}^{\text{VAT}}+ \lambda_{\text{SC}}\mathcal{L}_{\text{SC}}^{\text{VAT}},
	\end{align}
	where $\lambda_{\text{AR}}$ and $\lambda_{\text{SC}}$ are the loss weights to control the loss $\mathcal{L}_{\text{AR}}$ and $\mathcal{L}_{\text{SC}}$, respectively, in the AVT and VAT.
	Finally, we formulate the overall loss function of TransZero++:
	\begin{gather}
	\label{Eq:L_final}
	\begin{aligned}
	\mathcal{L}_{total}&=  \mathcal{L}_{\text{AVT}}+ \lambda_{\text{VAT}}\mathcal{L}_{\text{VAT}} \\
	& + \lambda_{\text{f\_SCL}}\mathcal{L}_{\text{f\_SCL}}+\lambda_{\text{p\_SCL}}\mathcal{L}_{\text{p\_SCL}},
	\end{aligned}
	\end{gather}
	where $\lambda_{\text{VAT}}$, $\lambda_{\text{f\_SCL}}$ and $\lambda_{\text{p\_SCL}}$ are the weights to control their corresponding loss terms. To enable the training process of TransZero++ more stable, we set the loss weight to one for $\mathcal{L}_{\text{AVT}}$.
	
	\begin{table*}[ht]
		\centering
		\caption{ Ablation studies for different components of TransZero++ on the CUB and SUN datasets. “FAE” is the feature augmentation encoder, “FA” means feature augmentation, and “DEC” denotes the decoders in AVT and VAT.} \label{table:ablation_component}
		\vspace{-2mm}
		\resizebox{0.9\linewidth}{!}{
			{
				\begin{tabular}{l|c|ccc|c|ccc}
					\hline
					\multirow{2}*{Method} &\multicolumn{4}{c|}{CUB} &\multicolumn{4}{c}{SUN}\\
					\cline{2-5}\cline{6-9}
					&\rm{acc}&\rm{U} & \rm{S} & \rm{H} &\rm{acc}&\rm{U} & \rm{S} & \rm{H}\\
					\hline
					\rowcolor{mygray}TransZero++ w/o AVT                                & 49.0& 36.4&48.5&41.6& 57.7& 36.9&28.7&32.3\\
					TransZero++ w/o VAT                                                  & 75.6& 62.8&72.3&67.2& 63.1& 44.1&35.5&39.3\\
					\rowcolor{mygray}TransZero++ w/o FAE                                 & 70.7& 63.4 &57.5 & 60.3& 64.0& 52.0&33.5&40.8\\
					TransZero++ w/o FA                                                   & 76.4& 66.6&70.3&68.4& 65.3& 46.7&36.9&41.2\\
					\rowcolor{mygray}TransZero++ w/o DEC                                 & 62.7& 49.2&63.4& 55.4& 64.1&48.4&34.0& 39.9\\
					TransZero++ (full)     &78.3&67.5&73.6&70.4&67.6&48.6&37.8&42.5\\
					\hline
			\end{tabular}}
		}
	\end{table*}
	
	\begin{table*}[ht]
		\centering
		\caption{ Ablation studies for different losses of TransZero++ on the CUB and SUN datasets. Note that $\mathcal{L}_{\text{SCL}}=\mathcal{L}_{\text{f\_SCL}}+\mathcal{L}_{\text{p\_SCL}}$.} \label{table:ablation_loss}
		\vspace{-2mm}
		\resizebox{0.9\linewidth}{!}{
			\begin{tabular}{l|c|ccc|c|ccc}
				\hline
				\multirow{2}*{Method} &\multicolumn{4}{c|}{CUB} &\multicolumn{4}{c}{SUN}\\
				\cline{2-5}\cline{6-9}
				&\rm{acc}&\rm{U} & \rm{S} & \rm{H} &\rm{acc}&\rm{U} & \rm{S} & \rm{H}\\
				\hline
				\rowcolor{mygray}TransZero++(VAT) w/o $\mathcal{L}_{\text{SCL}}$     & 49.0& 36.4&48.5&41.6& 57.7& 36.9&28.7&32.3\\
				TransZero++(AVT) w/o $\mathcal{L}_{\text{SCL}}$    & 75.6& 62.8&72.3&67.2& 63.1& 44.1&35.5&39.3\\
				\rowcolor{mygray}TransZero++(VAT) w/ $\mathcal{L}_{\text{SCL}}$     & 49.2&37.7&51.9 & 43.7& 63.3&48.0&31.5&38.0\\
				TransZero++(AVT) w/ $\mathcal{L}_{\text{SCL}}$     & 77.6& 67.2 & 73.4 & 70.2& 63.8&45.3&34.7&39.3\\
				\rowcolor{mygray}TransZero++ w/o $\mathcal{L}_{\text{SC}}$            & 77.0&46.6&76.4&58.9& 65.1&41.5&36.4&38.7\\
				TransZero++ w/o $\mathcal{L}_{\text{AR}}$            & 77.3& 67.1& 73.4&70.1& 64.7& 45.2&35.4&39.7\\
				\rowcolor{mygray}TransZero++(AVT and VAT) w/o $\mathcal{L}_{\text{f\_SCL}}$        & 78.1& 67.9&72.1&69.9& 65.6& 47.4&37.5&41.9\\
				TransZero++(AVT and VAT) w/o $\mathcal{L}_{\text{d\_SCL}}$        & 75.4& 64.4&71.0&67.5&65.2&46.0&37.6&41.4\\
				\rowcolor{mygray}TransZero (full)     &78.3&67.5&73.6&70.4&67.6&48.6&37.8&42.5\\
				\hline
			\end{tabular}
		}
	\end{table*}
	
	\subsection{Zero-Shot Prediction}
	After training TransZero++, We first obtain the visual embeddings of a test instance $x_i$ in the semantic space w.r.t. AVT and VAT, \textit{i.e.}, $\psi(x)$ and $\Psi(x)$. Considering the semantic-augmented visual embeddings learned by AVT and VAT are complementary to each other, we fuse their predictions using a combination coefficients $\alpha$ to predict the test label of $x_i$ with an explicit calibration, formulated as:
	\begin{gather}
	\label{eq:prediction}
	c^{*}=\arg \max _{c \in \mathcal{C}^u/\mathcal{C}}(\alpha\psi(x_i)+(1-\alpha)\Psi(x_i))^{\top} \times z^{c}+\mathbb {I}_{\left[c\in\mathcal{C}^u\right]}.
	\end{gather}
	Here, $\mathcal{C}^u/\mathcal{C}$ corresponds to the CZSL/GZSL setting, respectively. The complete procedures (including model training and prediction) for 
	TransZero++ are illustrated by the pseudocode in Algorithm \ref{algotithm:TransZero}.
	
	\section{Experiments}\label{sec4}
	\noindent\textbf{Dataset.} 	
	We conduct extensive experiments on four ZSL benchmarks, including two fine-grained datasets (\textit{i.e.}, CUB \cite{Welinder2010CaltechUCSDB2} and SUN \cite{Patterson2012SUNAD}), a coarse-grained dataset (\textit{i.e.}, AWA2 \cite{Xian2019ZeroShotLC}), and a large-scale dataset (\textit{i.e.}, ImageNet\cite{Deng2009ImageNetAL}). CUB has 11,788 images of 200 bird classes (seen/unseen classes = 150/50) depicted with 312 attributes. SUN includes 14,340 images from 717 scene classes (seen/unseen classes = 645/72) depicted with 102 attributes. AWA2 consists of 37,322 images from 50 animal classes (seen/unseen classes = 40/10) depicted with 85 attributes. We show some samples on these datasets in Fig. \ref{fig:samples}.

	\noindent\textbf{Evaluation Protocols.}
	Following \cite{Xian2019ZeroShotLC}, we measure the top-1 accuracy both in the CZSL and GZSL settings. In the CZSL setting, we predict the unseen classes to compute the accuracy of test samples, i.e., $\bm{acc}$. In the GZSL setting,  we calculate the accuracy of the test samples from both the seen classes (denoted as $\bm{S}$) and unseen classes (denoted as $\bm{U}$). Meanwhile, their harmonic mean (defined as $\bm{H =(2 \times S \times U) /(S+U)}$) is also used for evaluation in the GZSL setting.
	
	\noindent\textbf{Implementation Details.}
	We use the training splits proposed by \cite{Xian2019ZeroShotLC}. We take a ResNet101 pre-trained on ImageNet as the CNN backbone to extract the visual feature map $U^{\prime}(x)\in \mathbb{R}^{H \times W \times C}$ ($H$ and $W$ are the height and width of the feature maps, $C$ is the number of channels) without fine-tuning. We use the Adam optimizer with hyperparameters (learning rate = 0.001, betas = (0.5, 0.999)) to optimize our model. The learning rate and batch size are set to 0.0001 and 50 for all datasets, respectively. Following APN \cite{Xu2020AttributePN}, hyperparameters in our model are obtained by grid
		search on the validation set \cite{Xian2019ZeroShotLC}. Since the training data for the ZSL model is a medium scale which leads to over-fitting with more complex Transformer architectures, the encoder and decoder layers are set to 1 with one attention head both in AVT and VAT. We use PyTorch \cite{Paszke2019PyTorchAI} for the implementation of all experiments. More hyperparameter settings are shown in Sec. \ref{sec4.6}.

	\subsection{Comparison with State-of-the-Art}\label{sec4.1}	
	Our TransZero++ is a non end-to-end and non-generative manner. We compare it with other state-of-the-art methods both in CZSL and GZSL settings, including end-to-end methods (\textit{e.g.}, QFSL~\cite{Song2018TransductiveUE}, SGMA~\cite{Zhu2019SemanticGuidedML}, AREN~\cite{Xie2019AttentiveRE}, LFGAA~\cite{Liu2019AttributeAF}, APN~\cite{Xu2020AttributePN}, GEM-ZSL~\cite{Liu2021GoalOrientedGE}), generative methods (\textit{e.g.}, SE-ZSL~\cite{Arora2018GeneralizedZL}, f-VAEGAN~\cite{Xian2019FVAEGAND2AF}, OCD-CVAE~\cite{Keshari2020GeneralizedZL}, Composer~\cite{Huynh2020CompositionalZL}, E-PGN~\cite{Yu2020EpisodeBasedPG}, TF-VAEGAN~\cite{Narayan2020LatentEF}, IZF~\cite{Shen2020InvertibleZR}, SDGZSL~\cite{Chen2021SemanticsDF}, GCM-CF~\cite{Yue2021CounterfactualZA}, FREE~\cite{Chen2021FREE}, HSVA~\cite{Chen2021HSVA}, FREE+ESZSL~\cite{Cetin2022CL}) and non-generative methods (\textit{e.g.}, SP-AEN~~\cite{Chen2018ZeroShotVR}, PQZSL~\cite{Li2019CompressingUI}, IIR~~\cite{Cacheux2019ModelingIA}, TCN~\cite{Jiang2019TransferableCN}, DVBE~\cite{Min2020DomainAwareVB}, DAZLE~\cite{Huynh2020FineGrainedGZ}), to demonstrate its effectiveness and advantages.
	\subsubsection{Conventional Zero-Shot Learning}
	Here, we first compare our TransZero with the state-of-the-art methods in the CZSL setting. As shown in Table \ref{table:sota}, our TransZero++ achieves the best accuracy of 78.3\% and second-best accuracies of 67.6\%/64.6 on CUB and SUN/AWA2, respectively. This shows that TransZero++ effectively discovers the transferable semantic knowledge to represent the locality-augmented features, enabling desirable knowledge transfer for distinguishing various unseen classes. Compared with other attention-based methods (\textit{e.g.}, SGMA~\cite{Zhu2019SemanticGuidedML}, AREN~\cite{Xie2019AttentiveRE}, APN~\cite{Xu2020AttributePN}), TransZero++ gets significant gains over 6.3\% and 5.0\% on CUB and SUN, respectively. This demonstrates that the attribute localization representations learned by our TransZero++ are more transferable than the region embeddings learned by the existing attention-based methods on fine-grained datasets. Because TransZero++ represents the key transferable semantic knowledge with high confidence, which significantly suppresses the common knowledge between various fine-grained classes.
	Benefiting from the semantic collaborative learning and the two complementary semantic-augmented embeddings learned by AVT and VAT, TransZero++ further improves the performance over its conference version (TransZero \cite{Chen2021TransZero}).

	To further validate the effectiveness of TransZero++, we compare it with existing  methods on large-scale dataset (\textit{i.e.}, ImageNet). To facilitate fair comparison, we follow the train/test split suggested in \cite{Kampffmeyer2019RethinkingKG}\footnote{https://github.com/yinboc/DGP}. Table \ref{table:ImageNet} shows the top-1 accuracy of various methods in the “2-hops” setting, which contains all the classes within two hops from the seen classes\cite{Xian2019ZeroShotLC,Kampffmeyer2019RethinkingKG}. We replace attribute features with w2v for TransZero++. Results show that TransZero++ outperforms existing ZSL models. This indicates TransZero++ has potential advantages in ZSL via semantic collaborative learning. Because SGCN\cite{Kampffmeyer2019RethinkingKG} employs a graph convolutional neural network (GCN) to discover the prior information of hierarchical relationships among various classes, it achieves better results than our TransZero++.

	
	\subsubsection{Generalized Zero-Shot Learning}

	Table \ref{table:sota} shows the results of different methods in the GZSL setting. We can see that most state-of-the-art methods achieve good results on seen classes but fail on unseen classes, while our TransZero++ generalizes better to unseen classes with high unseen and seen accuracies.  For example, TransZero++ obtains the best performance with a harmonic mean of 70.4\% and 72.5\% on CUB and AWA2, respectively. We argue these desirable results benefit from the fact that i) the feature augmentation encoders in AVT and VAT effectively refine the visual features that are more discriminative and transferable than the ones directly extracted from the CNN backbone; ii) the VAT and AVT discover the key semantic knowledge between visual and attribute features for locality-augmented feature representations, enabling effective knowledge transfer from seen to unseen classes. Since Transzero++ is an embedding-based method, it cannot achieve best results on SUN compared to the strong generative methods, \textit{e.g.}, OCD-CVAE~\cite{Keshari2020GeneralizedZL}, TF-VAEGAN~\cite{Narayan2020LatentEF}, HSVA~\cite{Chen2021HSVA} and IZF~\cite{Shen2020InvertibleZR}. Since per class only contains about 16 training images on SUN, which heavily limits the ZSL models, the data augmentation is very effective for improving the performance on SUN. Benefiting from that APN+f-VAEGAN-D2 extractes the discirminate visual features to learn a good transductive generative model (e.g., f-VAEGAN-D2) for data augmentation, it achieves the best performance on SUN. As such, most of the strong generative methods perform better than our TransZero++ (embedding-based method) on SUN. Compared to the latest attention-based method (\textit{e.g.}, APN \cite{Xu2020AttributePN}), our TransZero++ achieves significant improvements of 3.2\%, 4.9\% and 9.6\% in harmonic mean on CUB, SUN and AWA2, respectively. This demonstrates the superiority and great potential of our cross attribute-guided Transformer for the ZSL task. Since the semantical collaborative learning encourages AVT and VAT to learn semantic-augmented embedding for desirable visual-semantic interaction, TransZero++ continuously improves the performance of its conference version (TransZero \cite{Chen2021TransZero}) on all datasets.

	\begin{table*}[ht]
		\centering
		\caption{ Ablation studies for different components of TransZero++ on the CUB and SUN datasets. “FAE” is the feature augmentation encoder, “FA” means feature augmentation, and “DEC” denotes visual-semantic decoder.} \label{table:ablation_metric}
		\vspace{-2mm}
		\resizebox{0.9\linewidth}{!}{
			{
				\begin{tabular}{l|c|ccc|c|ccc}
					\hline
					\multirow{2}*{Method} &\multicolumn{4}{c|}{CUB} &\multicolumn{4}{c}{SUN}\\
					\cline{2-5}\cline{6-9}
					&\rm{acc}&\rm{U} & \rm{S} & \rm{H} &\rm{acc}&\rm{U} & \rm{S} & \rm{H}\\
					\hline
					\rowcolor{mygray}TransZero++ w/ $\ell_1$                           &53.6&42.6&48.9&37.8&64.3&51.4&32.7&40.0\\
					TransZero++ w/ KL($p_1||p_2$)                              &76.4& 67.8&65.4&70.4&65.0&46.3&37.6&41.5\\
					\rowcolor{mygray}TransZero++ w/ KL($p_2||p_1$)              &76.3& 68.3&65.3&71.6&65.0&46.3&38.1&41.8\\
					TransZero++ w/ JSD                                                &76.1&67.2&63.2&71.8&64.8&46.2&37.9&41.6\\
					\rowcolor{mygray}TransZero++ w/ $\ell_2$           &78.3&67.5&73.6&70.4&67.6&48.6&37.8&42.5\\
					\hline
			\end{tabular}}
		}
	\end{table*}

	\subsection{Ablation Study}\label{sec4.2}
	To provide further insight into TransZero++, we conduct ablation studies to evaluate the effect of different model components, loss functions, and distance metrics for semantical collaborative losses.
	
	\noindent\textbf{Analysis of Model Components.}As shown in Table \ref{table:ablation_component}, we conduct ablation studies to evaluate the effects of different model components , \textit{i.e.}, AVT, VAT, feature augmentation encoder (denoted as FAE), feature augmentation in FAE (denoted as FA), and visual-semantic decoder (denoted as DEC).  We observed that TransZero++ with various model components achieve more clear improvements on CUB than on SUN, because SUN is a scene dataset, which is more complicated and challenging than CUB. TransZero++ performs significantly worse than if no AVT is used, \textit{i.e.}, the acc/harmonic mean drops by 29.3\%/28.8\% on CUB and 9.9\%/10.2\% on SUN. This indicates that AVT is an basic attention sub-net in TransZero++, which is consistent with most existing attention-based methods \cite{Huynh2020FineGrainedGZ,Xu2020AttributePN,Liu2021GoalOrientedGE,Chen2022MSDN} based on attribute$\rightarrow$visual attention. Meanwhile, Transzero++ w/o VAT also achieves inferior performance than the full model.  As such, it is necessary to simultaneously learn the semantic-augmented embeddings with AVT and VAT for ZSL, under the guidance of semantical collaborative learning. TransZero++ significantly improves its performance when AVT and VAT use the feature augmentation encoders, which shows the importance of refining the visual feature to alleviate the cross-dataset bias. If we incorporate the encoder of the standard Transformer without feature augmentation, TransZero++ obtains inferior performances as the entangled relative geometry priors limit the transferability of visual features, \textit{i.e.}, the acc/harmonic mean drops by 1.9\%/2.0\% and 2.3\%/1.3\% on CUB and SUN, respectively. When TransZero++ does not employ the decoders in AVT and VAT to localize the key object attribute for semantic knowledge representations, its performance decreases dramatically on all datasets. 
	
	\noindent\textbf{Analysis of Loss Functions.} As shown in Table \ref{table:ablation_loss}, we further conduct ablation studies to evaluate the effects of different loss functions, \textit{i.e.}, semantical collaborative loss (including $\mathcal{L}_{\text{f\_SCL}}$ and $\mathcal{L}_{\text{p\_SCL}}$), self-calibration loss (\textit{i.e.}, $\mathcal{L}_{\text{SC}}$) and attribute regression loss (\textit{i.e.},  $\mathcal{L}_{\text{AR}}$). TransZero++ using single sub-net (\textit{i.e.}, TransZero++(VAT) and TransZero++(VAT)) achieves significant gains on CUB and SUN when it uses the semantical collaborative loss ($\mathcal{L}_{\text{SCL}}=\mathcal{L}_{\text{f\_SCL}}+\mathcal{L}_{\text{p\_SCL}}$). For example, TransZero++(VAT) and TransZero++(AVT) achieve gains of 2.1\% and 3.0\% in harmonic mean on CUB, respectively. This shows that semantic collaborative learning is effective for encouraging AVT and VAT to teach each other to discover the key transferable semantic knowledge for ZSL. The self-calibration mechanism can effectively alleviate the seen-unseen bias problem \cite{Zhu2019SemanticGuidedML,Huynh2020FineGrainedGZ,Xu2020AttributePN}, resulting in improvements in the harmonic mean of 11.5\% on CUB. The attribute regression constraint further improves the performance of TransZero++ by directing $\mathcal{M}_1$ and $\mathcal{M}_2$ to conduct effective visual-semantic mapping. Furthemore, the two semantical collaborative losses (\textit{i.e.}, $\mathcal{L}_{\text{f\_SCL}}$ and $\mathcal{L}_{\text{p\_SCL}}$) encourage TransZero++ to conduct desirable semantical collaborative learning.
	
	\noindent\textbf{Analysis of Distance Metrics for Semantical Collaborative Losses.}  As shown in Table \ref{table:ablation_metric}, we conduct ablation studies to evaluate the effects of distance metrics for semantical collaborative losses (feature-level semantical collaborative loss ($\mathcal{L}_{\text{f\_SCL}}$) and prediction-level semantical collaborative loss ($\mathcal{L}_{\text{p\_SCL}}$)), \textit{i.e.}, $\ell_1$, $\ell_2$, KL($p_1||p_2)$, KL($p_2||p_1)$, and JSD.  Results show that TransZero++ performs very poorly using $\ell_1$ distance for calculating the semantical collaborative losses. The possible reason is that the values of semantic-augmented visual embeddings and predictions in VAT and AVT are constrained to be in a small range, $\ell_1$ distance cannot well capture the divergence between the outputs of AVT and VAT.   When TransZero++ use the KL($p_1||p_2$), KL($p_2||p_1$), or JSD to compute the semantical collaborative losses, it achieves consistent good performance almost. Thus, the symmetric and asymmetric distances for semantical collaborative losses do not make any difference. Interestingly, TransZero++ achieves the best results using $\ell_2$ distance which is beneficial for the regression problem. As such, we take $\ell_2$ to compute $\mathcal{L}_{\text{p\_SCL}}$ and $\mathcal{L}_{\text{f\_SCL}}$.
	
	\begin{figure*}[t]
		\begin{center}
			\hspace{0.5mm}\rotatebox{90}{\hspace{0.4cm}{\footnotesize (a) AREN }}\hspace{-1mm}
			\includegraphics[width=17cm,height=2.05cm]{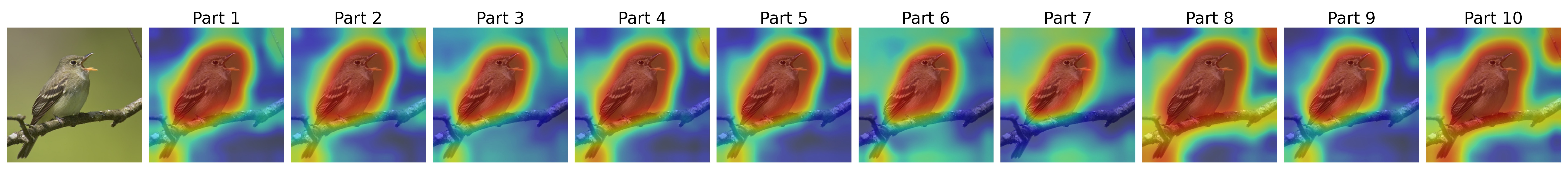}\\
			\hspace{0.5mm}\rotatebox{90}{\hspace{0.3cm}{\footnotesize (b) TransZero }}\hspace{-1mm}
			\includegraphics[width=17cm,height=2.55cm]{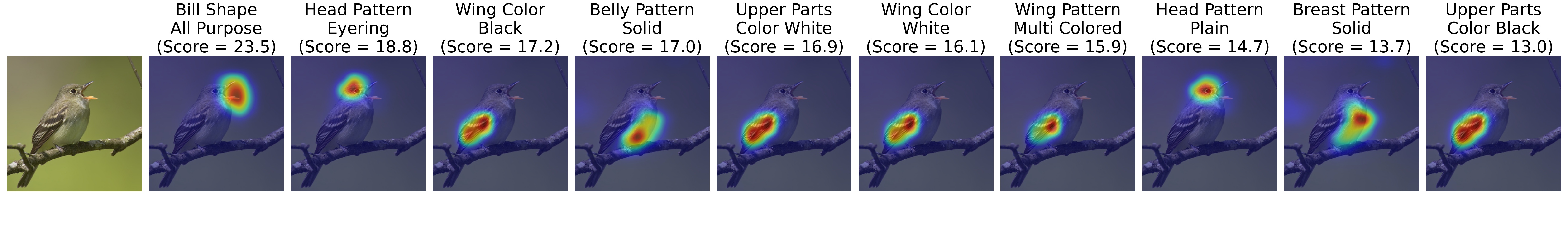}\\\vspace{-3mm}
			\hspace{0.5mm}\rotatebox{90}{\hspace{0.3cm}{\footnotesize (c) TransZero++ }}\hspace{-1mm}
			\includegraphics[width=17cm,height=2.55cm]{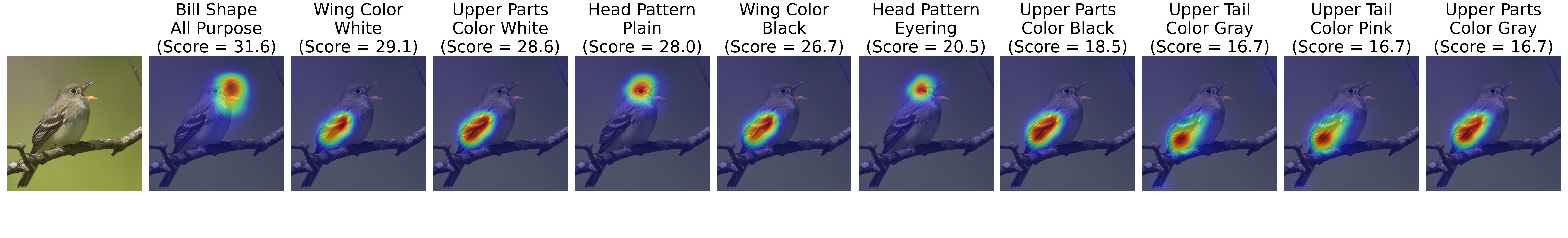}\\\vspace{-3mm}
			\caption{Visualization of top-10 attention maps for the attention-based method (\textit{i.e.}, AREN \cite{Xie2019AttentiveRE}), TransZero \cite{Chen2021TransZero} (Conference version) and our TransZero++. Results show that TransZero localizes some important object attributes with low confidence scores for representing region features, while AREN is failed. Furthermore, our TransZero++ discovers more valuable attributes that exist in the corresponding image with high confidence scores compared to the TransZero. More results are presented in the \href{https://shiming-chen.github.io/TransZero-pp/TransZero-pp.html}{Project Website}. (Best viewed in color)}
			\label{fig:attened-part}
		\end{center}\vspace{-6mm}
	\end{figure*}
	
	\begin{figure*}[t]
		\begin{center}
			\hspace{0.5mm}\rotatebox{90}{\hspace{2.7cm}{\footnotesize (a) Seen Classes }}\hspace{0mm}
			\includegraphics[width=16cm,height=8cm]{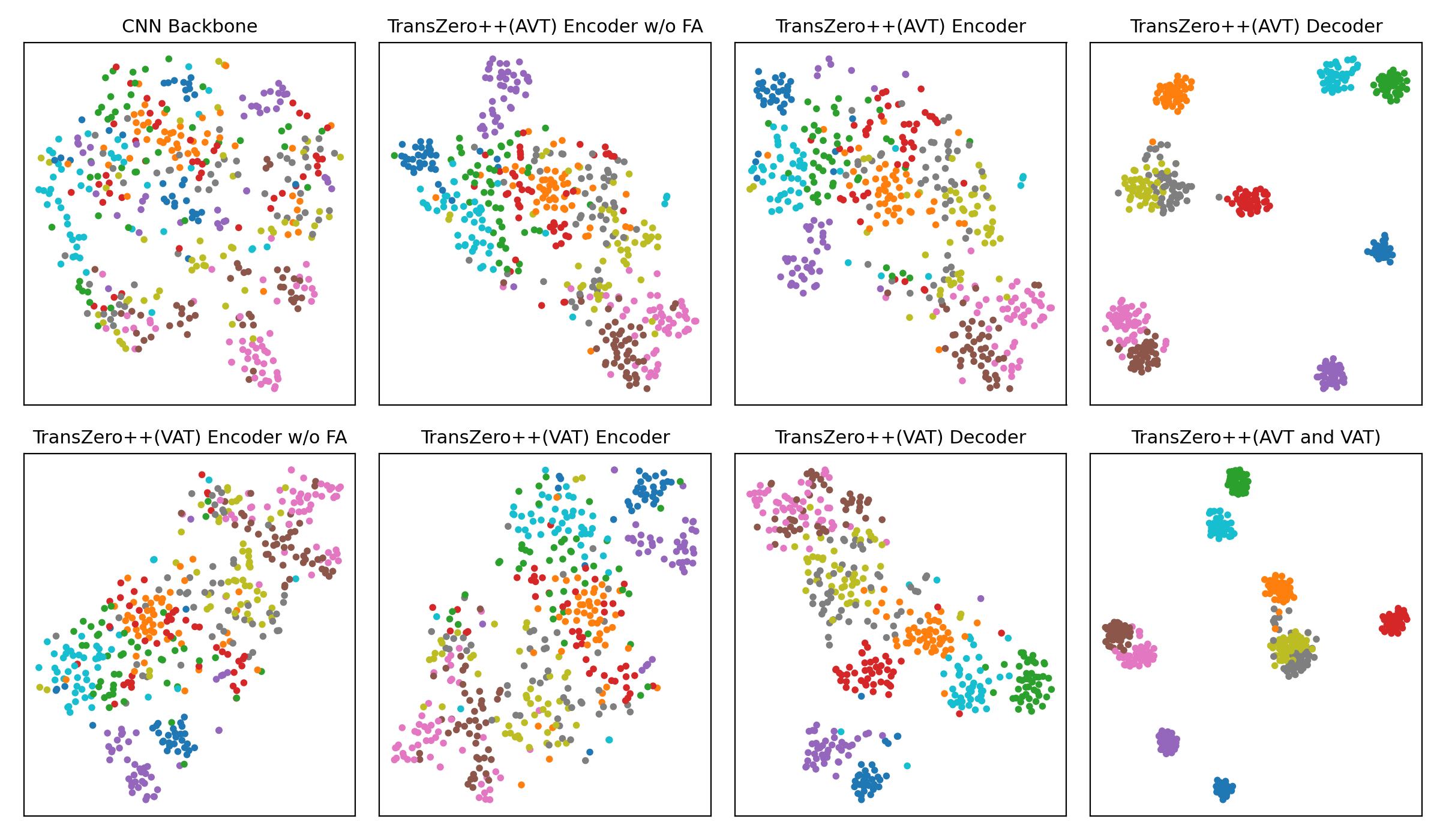}\\ 
			\hspace{0.5mm}\rotatebox{90}{\hspace{2.6cm}{\footnotesize (b) Unseen Classes }}\hspace{0.7mm}
			\includegraphics[width=16cm,height=8cm]{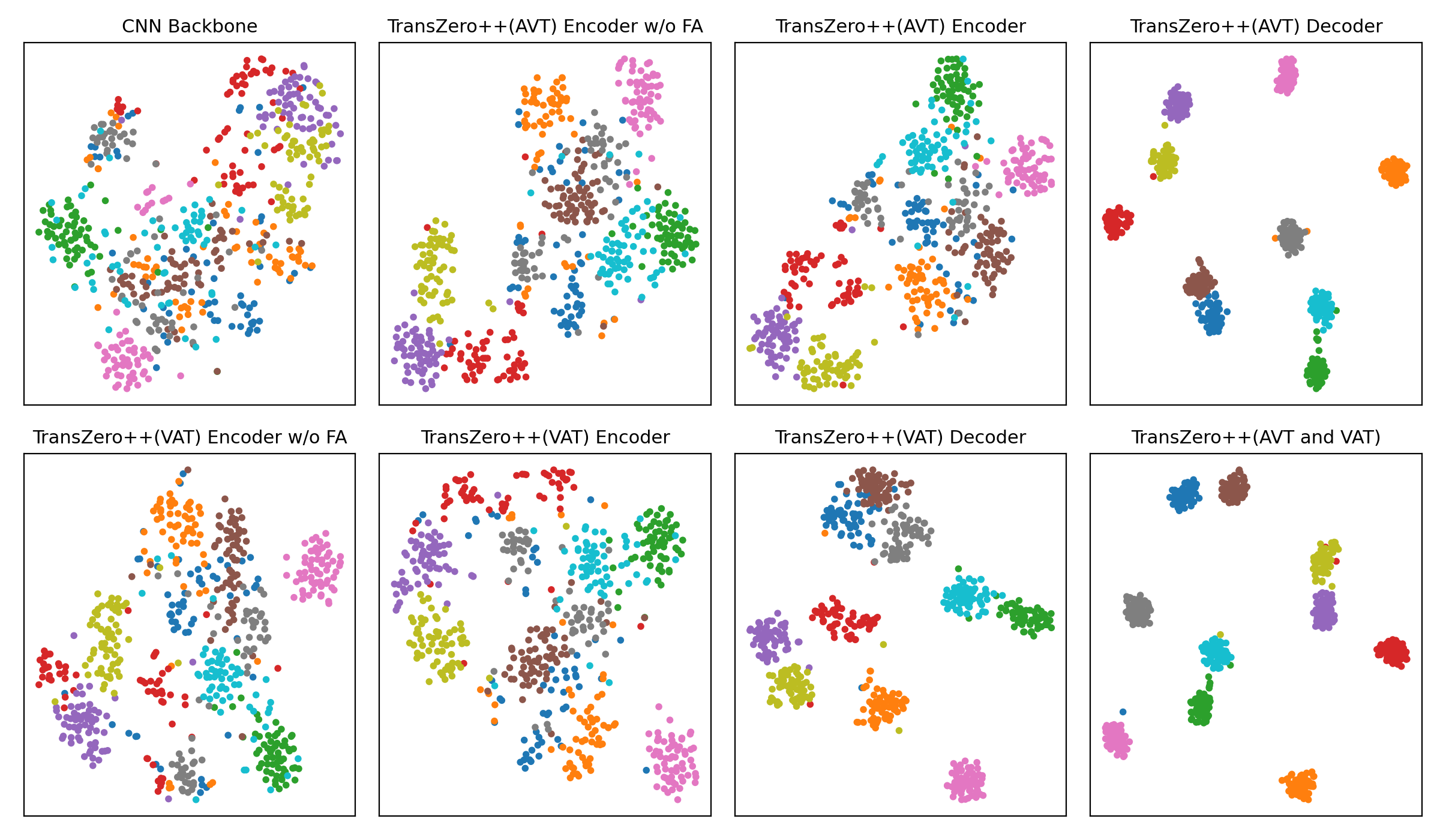}\\ \vspace{-2mm}
			\caption{t-SNE visualizations of visual features for (a) seen classes and (b) unseen classes, learned by the CNN backbone, TransZero++(AVT) encoder w/o FA, TransZero++(AVT) encoder, TransZero++(AVT) decoder, TransZero++(VAT) encoder w/o FA, TransZero++(VAT) encoder, TransZero++(VAT) decoder and TransZero++(VAT and VAT). The 10 colors denote 10 different seen/unseen classes randomly selected from CUB. Results show that our various model components in TransZero++ learn the discriminative visual feature representations, while CNN backbone (\textit{e.g.}, ResNet101) failed. The results on SUN and AWA2 are presented in the \href{https://shiming-chen.github.io/TransZero-pp/TransZero-pp.html}{Project Website}. (Best viewed in color)}
			\label{fig:tsne}
		\end{center}\vspace{-6mm}
	\end{figure*}

	\subsection{Qualitative Results}\label{sec4.3}	
	Here, we present the visualizations of attention maps and t-SNE \cite{Maaten2008VisualizingDU} to intuitively show the effectiveness of our TransZero++.
	\subsubsection{Visualization of Attention Maps.} To intuitively show the effectiveness of our TransZero++ at learning attribute-relevant visual features, we visualize the attention maps learned by the existing attention-based methods (\textit{e.g.}, AREN \cite{Xie2019AttentiveRE}) and TransZero++. As shown in Fig. \ref{fig:attened-part}, AREN simply learns region embeddings for visual representations, \textit{e.g.}, the whole bird body, neglecting the fine-grained semantic attribute information. In contrast, our Transzero++ learns discriminative attribute localization for visual features by assigning high positive scores to key attributes (\textit{e.g.}, the “bill shape all-purpose” of the \textit{Acadian Flycatcher} in Fig. \ref{fig:attened-part}). Thus, TransZero++ discovers the semantic-augmented embeddings that are discriminative and transferable, enabling good performance both in seen and unseen classes. Compared to TransZero \cite{Chen2021TransZero} (Conference version), TransZero++ can discover more valuable attributes for semantic-augmented embedding representations (\textit{e.g.}, “upper part color gray” of \textit{Acadian Flycatcher}). Furthermore, TransZero++ gets higher confidence scores for the important attributes that exist in the images than TransZero. For example, TransZero++ gets the scores of 28.0 for attribute “head pattern plain” of the \textit{Acadian Flycatcher}, while TransZero gets the scores of 14.7.

	\begin{figure*}[t]
		\begin{center}
			\hspace{0.5mm}\rotatebox{90}{\hspace{1.0cm}{\footnotesize (a) CUB}}\hspace{0mm}
			\hspace{-0mm}\includegraphics[width=3.7cm,height=2.8cm]{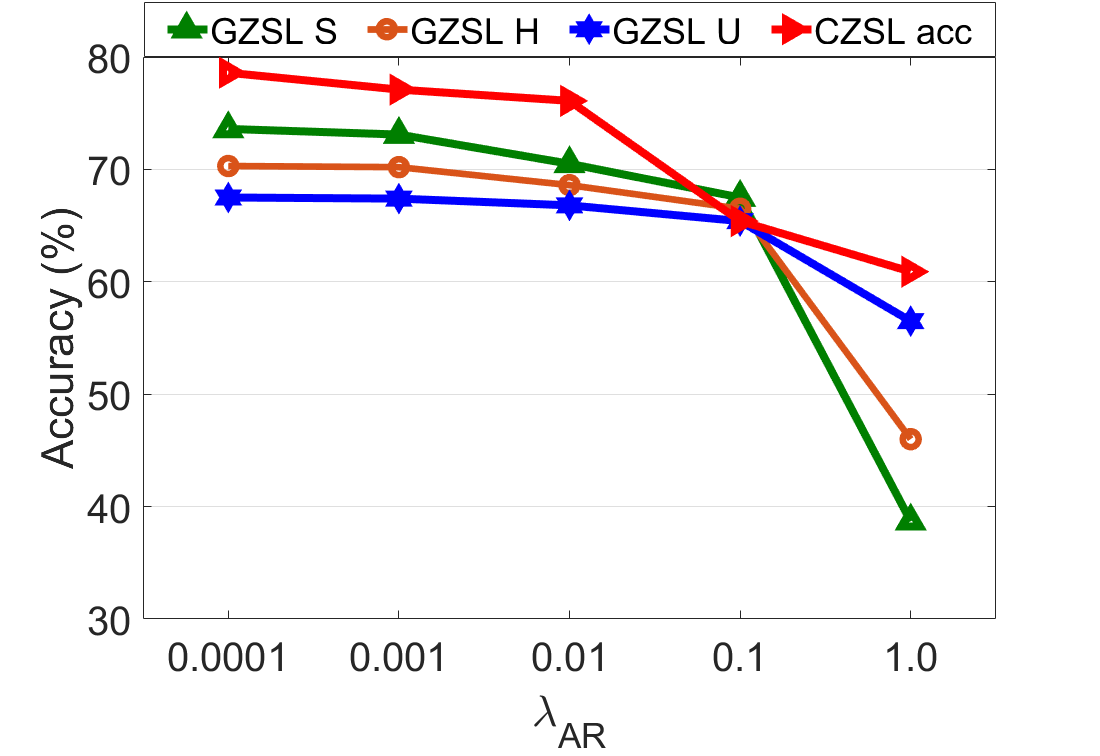}
			\hspace{-3mm}\includegraphics[width=3.7cm,height=2.8cm]{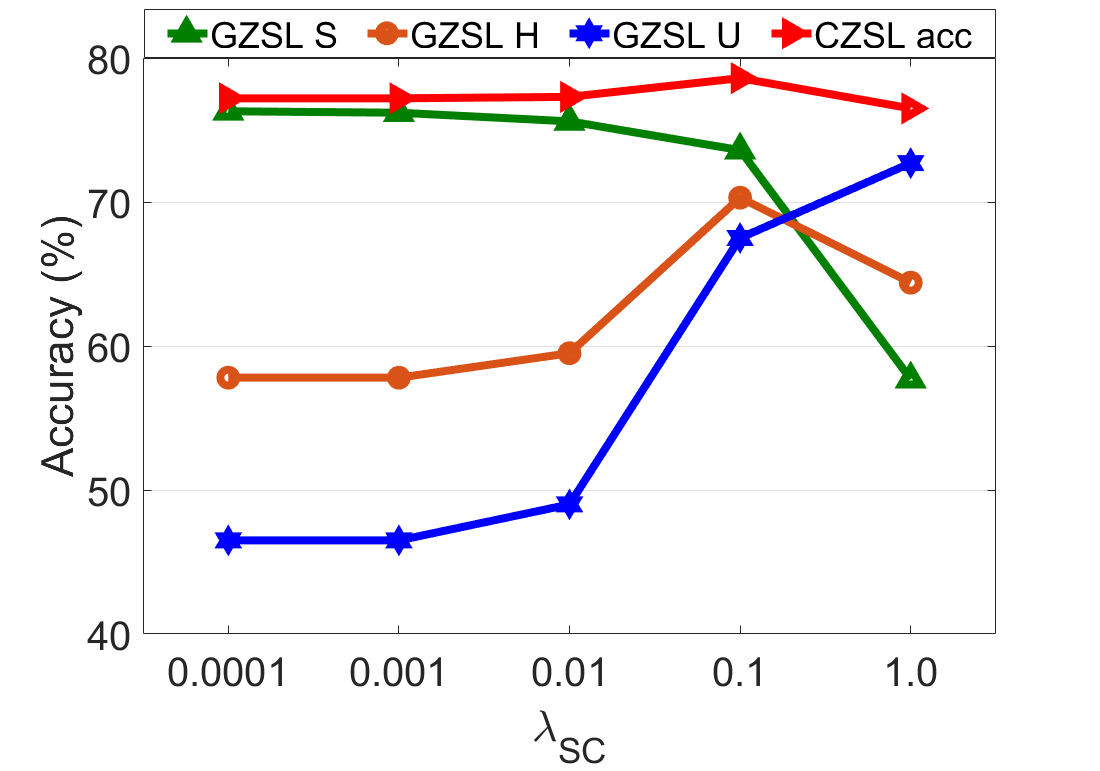}
			\hspace{-3mm}\includegraphics[width=3.7cm,height=2.8cm]{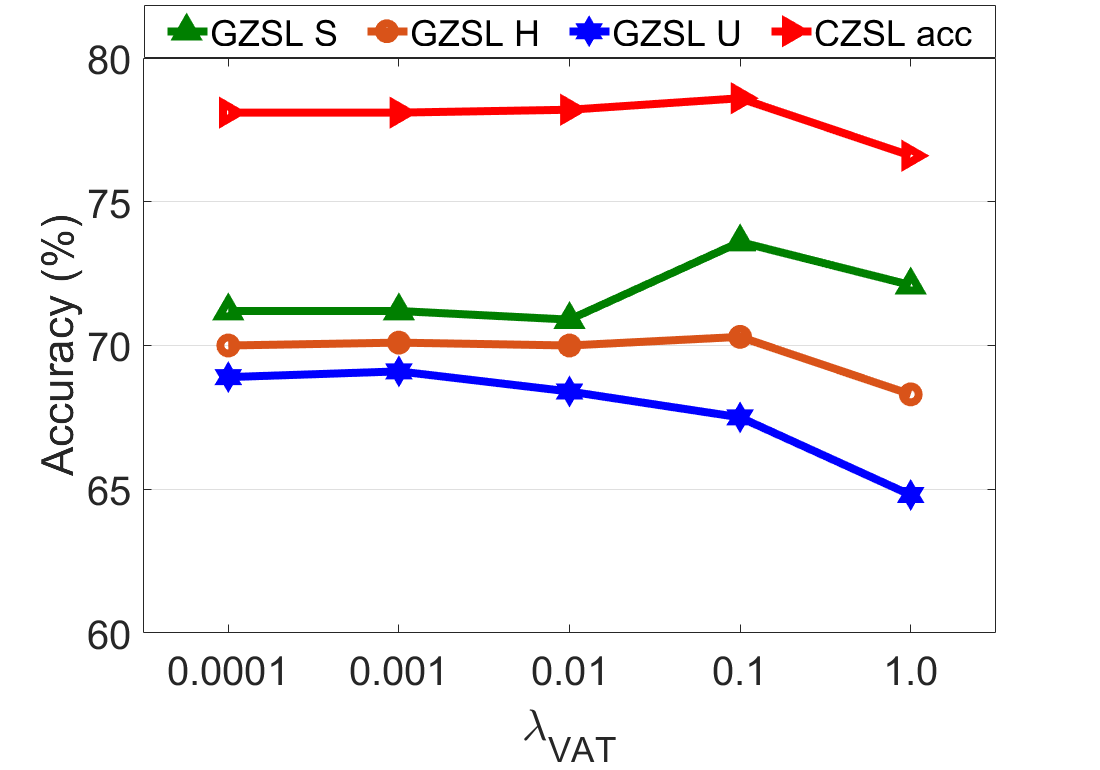}
			\hspace{-3mm}\includegraphics[width=3.7cm,height=2.8cm]{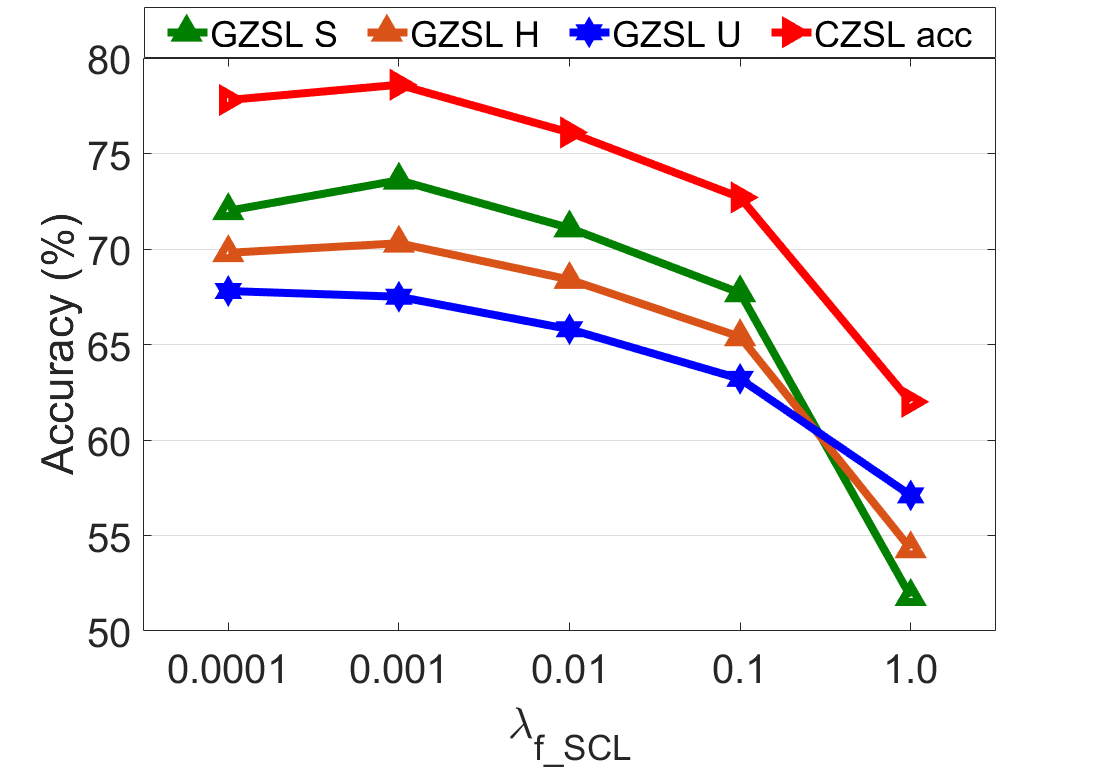}
			\hspace{-3mm}\includegraphics[width=3.7cm,height=2.8cm]{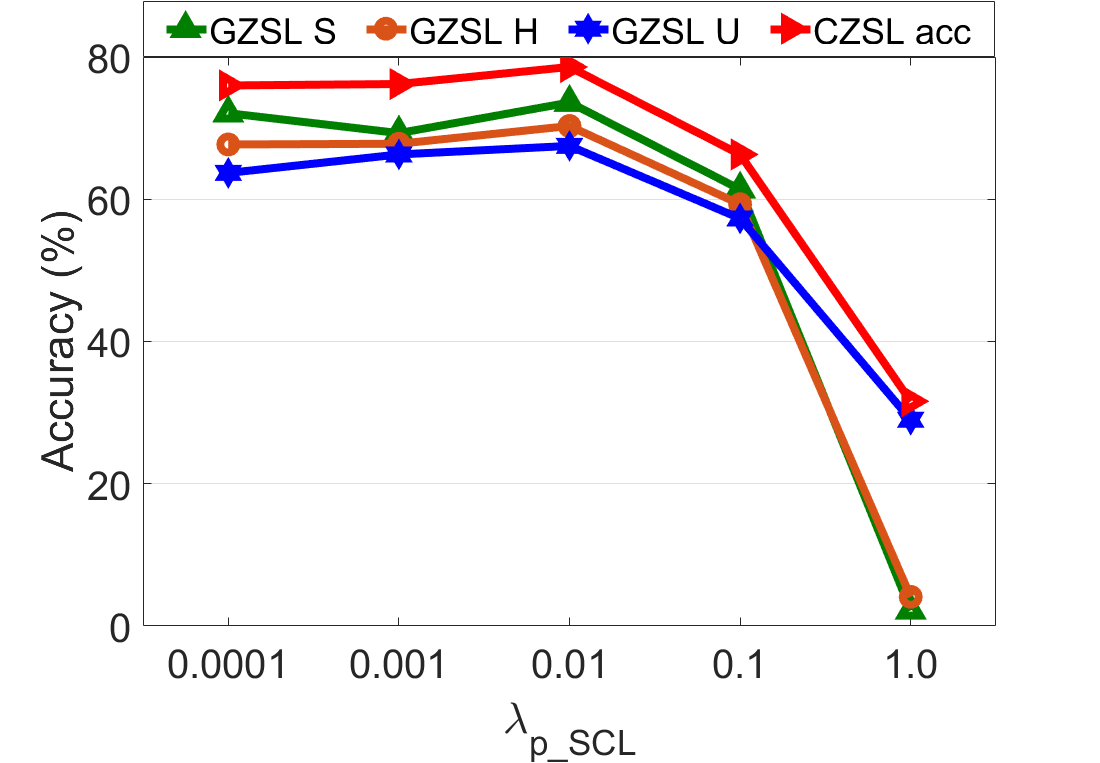}\\\vspace{2mm}
			\hspace{0.5mm}\rotatebox{90}{\hspace{1.0cm}{\footnotesize (b) SUN}}\hspace{0mm}
			\hspace{-0mm}\includegraphics[width=3.7cm,height=2.8cm]{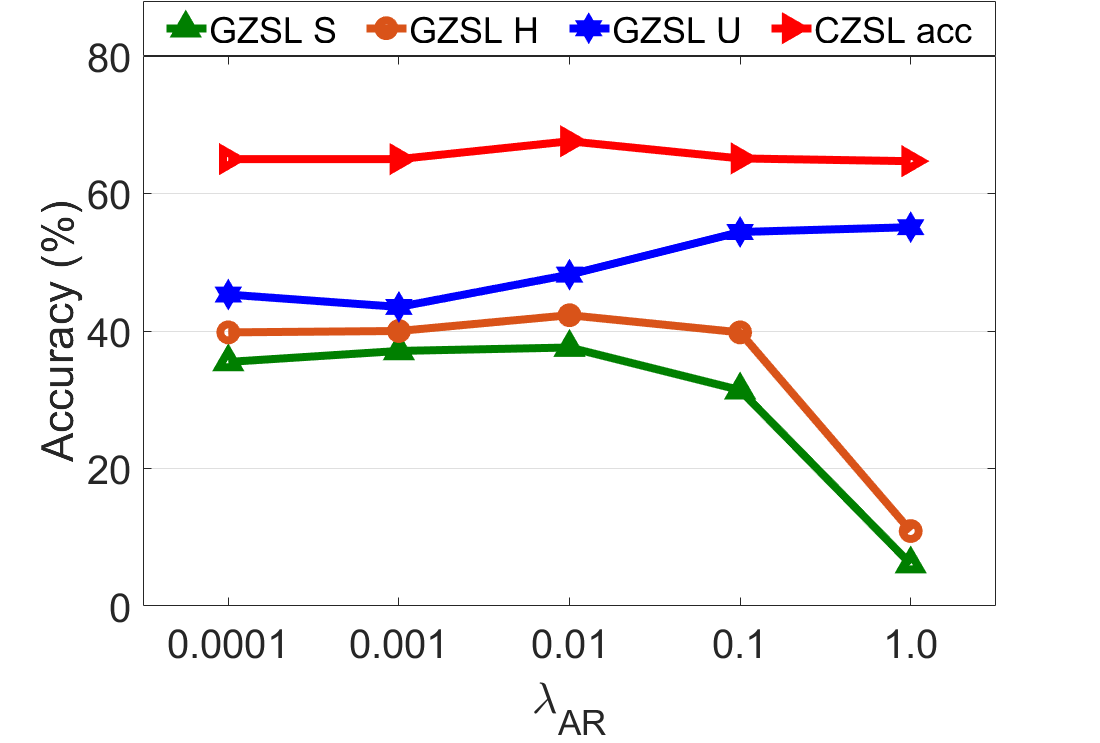}
			\hspace{-3mm}\includegraphics[width=3.7cm,height=2.8cm]{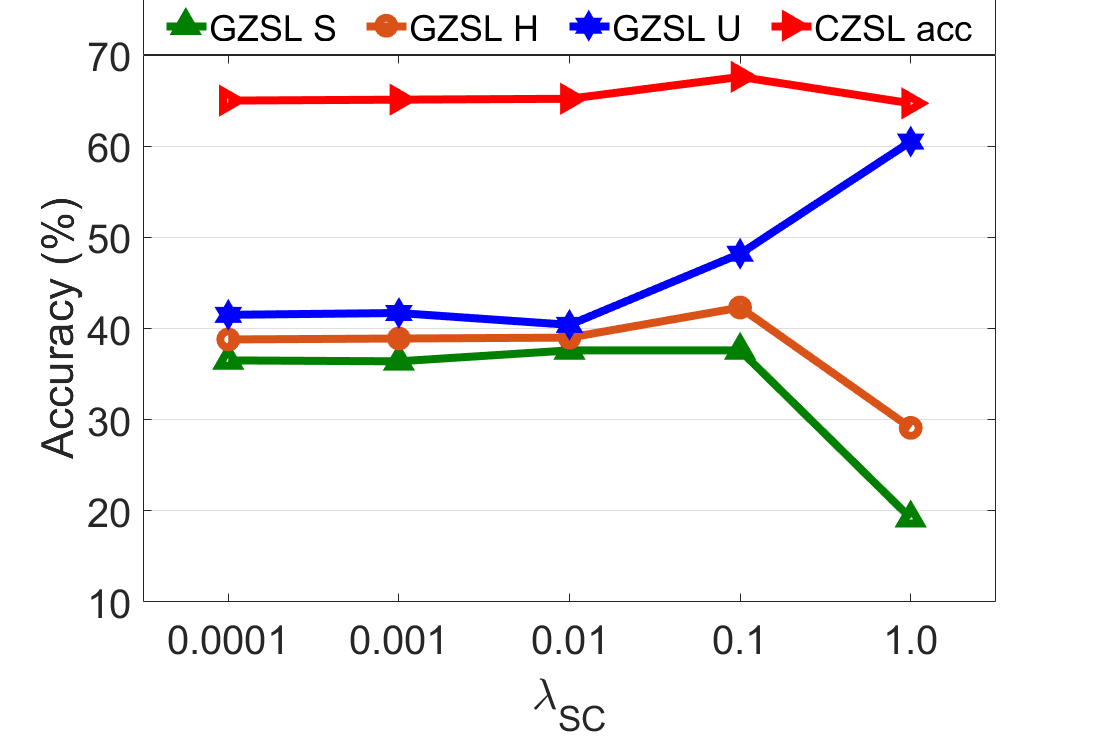}
			\hspace{-3mm}\includegraphics[width=3.7cm,height=2.8cm]{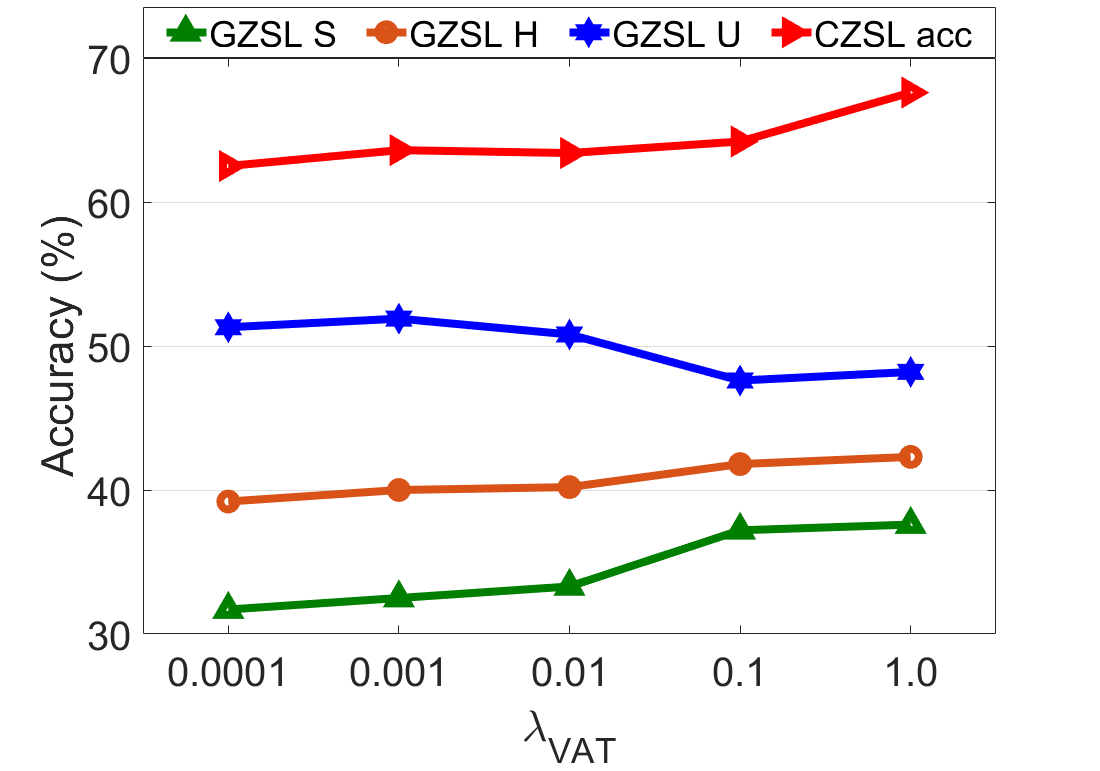}
			\hspace{-3mm}\includegraphics[width=3.7cm,height=2.8cm]{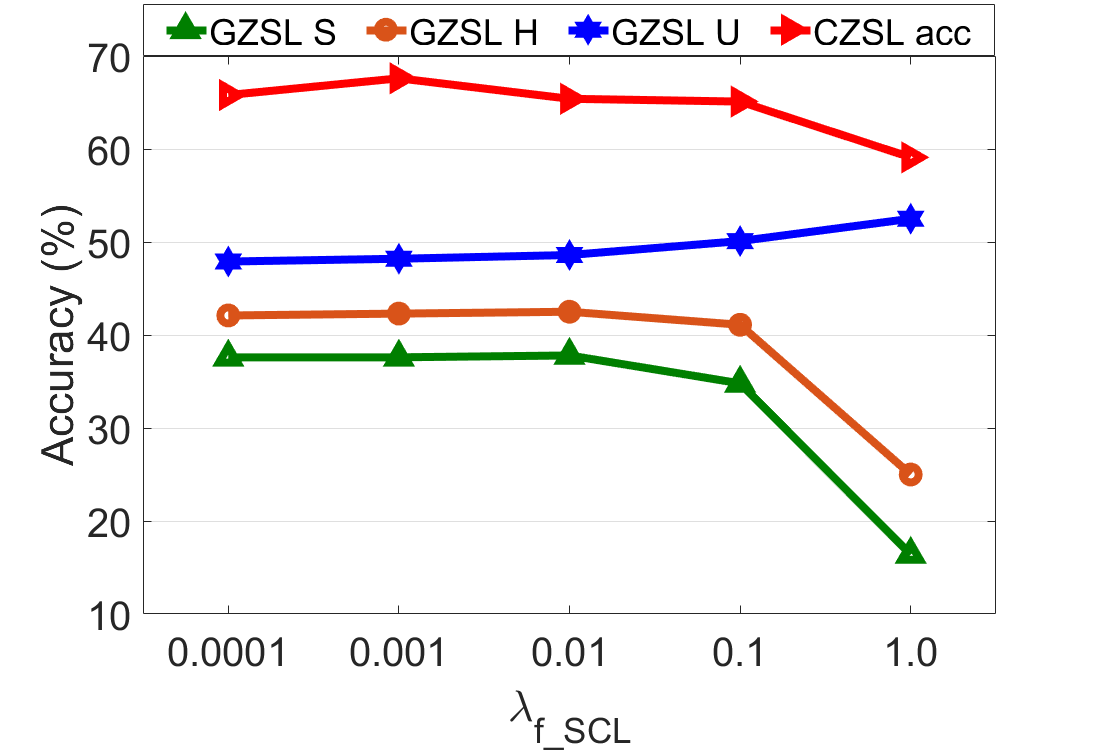}
			\hspace{-3mm}\includegraphics[width=3.7cm,height=2.8cm]{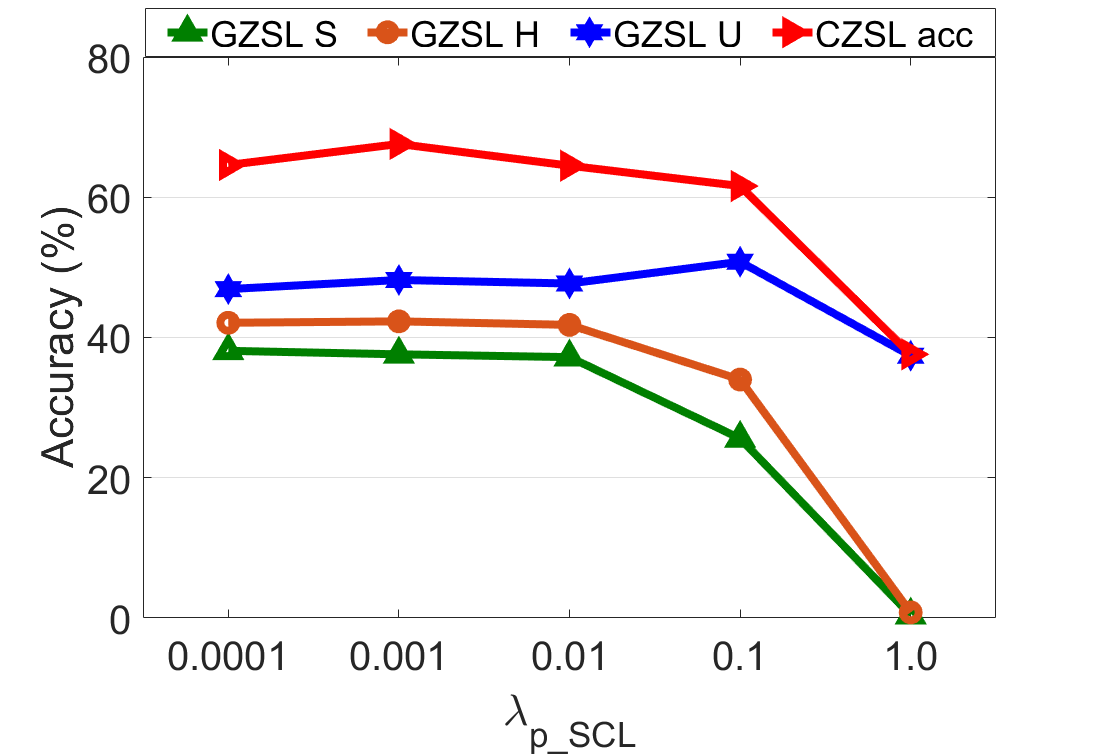}\\
			\caption{The effects of loss weights that control their corresponding loss terms on CUB and SUN, \textit{i.e.},  $\lambda_{\text{AR}}$, $\lambda_{\text{SC}}$, $\lambda_{\text{VAT}}$, $\lambda_{\text{f\_SCL}}$ and $\lambda_{\text{p\_SCL}}$.}  
			\label{fig:loss-weights}
		\end{center}\vspace{-6mm}
	\end{figure*}

	\subsubsection{t-SNE Visualizations.} As shown in Fig. \ref{fig:tsne}, we provide the t-SNE visualization \cite{Maaten2008VisualizingDU} of visual features for (a) seen classes and (b) unseen classes on CUB, learned by the CNN backbone, TransZero++(AVT) encoder w/o FA, TransZero++(AVT) encoder, TransZero++(AVT) decoder,  TransZero++(VAT) encoder w/o FA, TransZero++(VAT) encoder, TransZero++(VAT) decoder, TransZero++(AVT and VAT). If the standard encoder is incorporated into AVT and VAT of our TransZero++, the visual features learned by the encoder are significantly improved compared to the original visual features extracted from the CNN Backbone (\textit{e.g.}, ResNet101). When we use the feature augmentation encoder to refine the original visual features, the quality of visual features is further enhanced. These results demonstrate that the encoder of TransZero++ effectively alleviates the cross-dataset bias problem and reduces the entangled relative geometry relationships among different regions, enabling the visual feature to be more discriminative and transferable. Moreover, the attribute$\rightarrow$visual and visual$\rightarrow$attribute decoders in AVT and VAT learn attribute-based visual features and visual-based attribute features, which are further mapped into semantic embedding space for semantic-augmented embedding representations. Since the features learned by AVT and VAT are complementary to each other, the fused semantic-augmented embedding can be further refined. As such, our TransZero++ achieves significant performance both in seen and unseen classes on all datasets.

	\begin{figure}[t]
		\begin{center}
			\hspace{0.5mm}\rotatebox{90}{\hspace{1.3cm}{\footnotesize (a) CUB }}\hspace{0mm}
			\includegraphics[width=4.1cm,height=3.2cm]{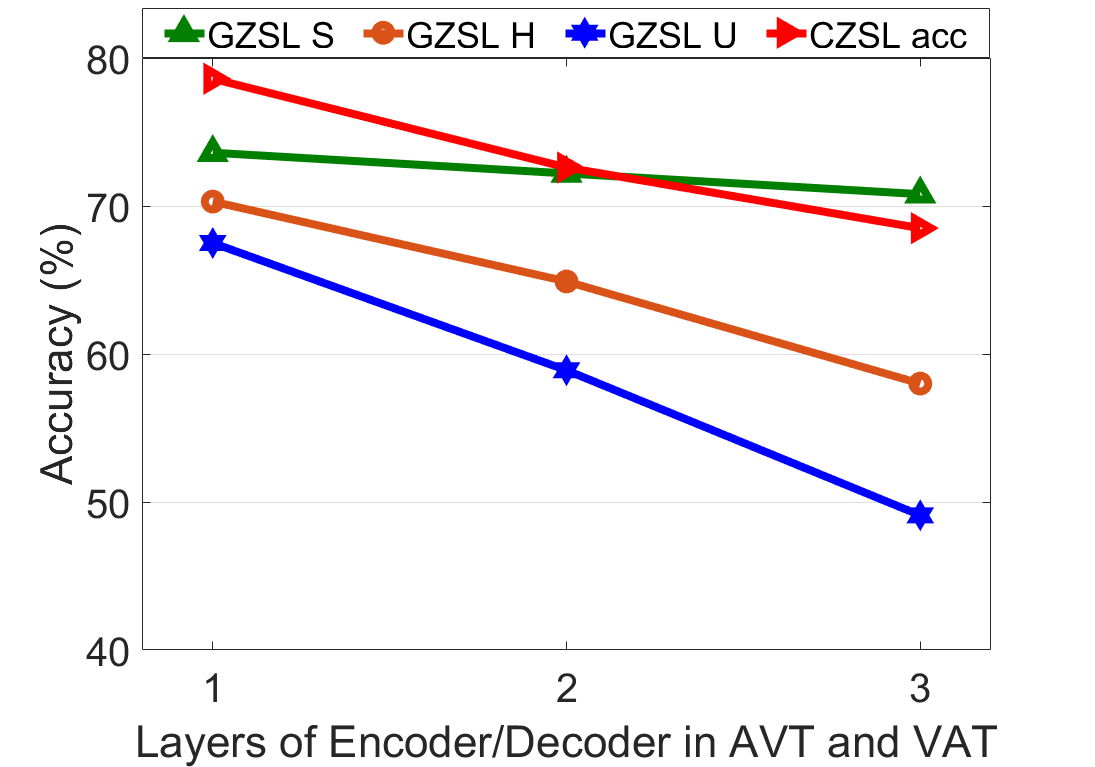}
			\includegraphics[width=4.1cm,height=3.2cm]{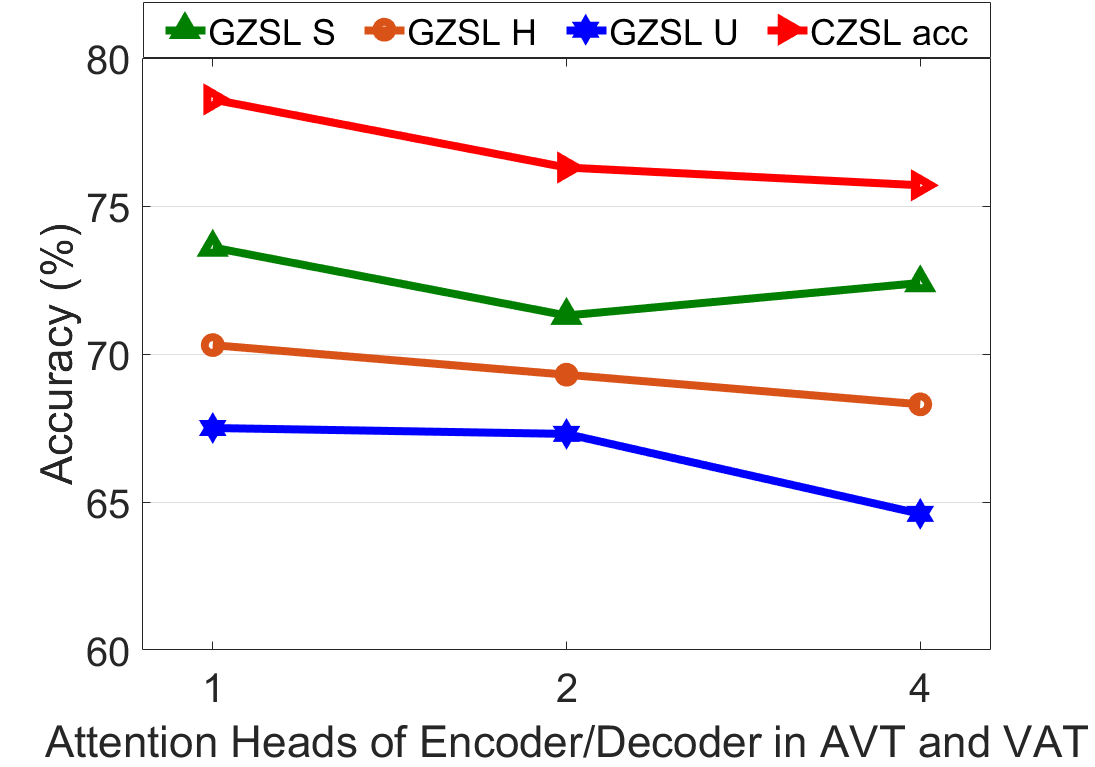}\vspace{2mm}
			\hspace{0.5mm}\rotatebox{90}{\hspace{1.3cm}{\footnotesize (b) SUN }}\hspace{0mm}
			\includegraphics[width=4.1cm,height=3.2cm]{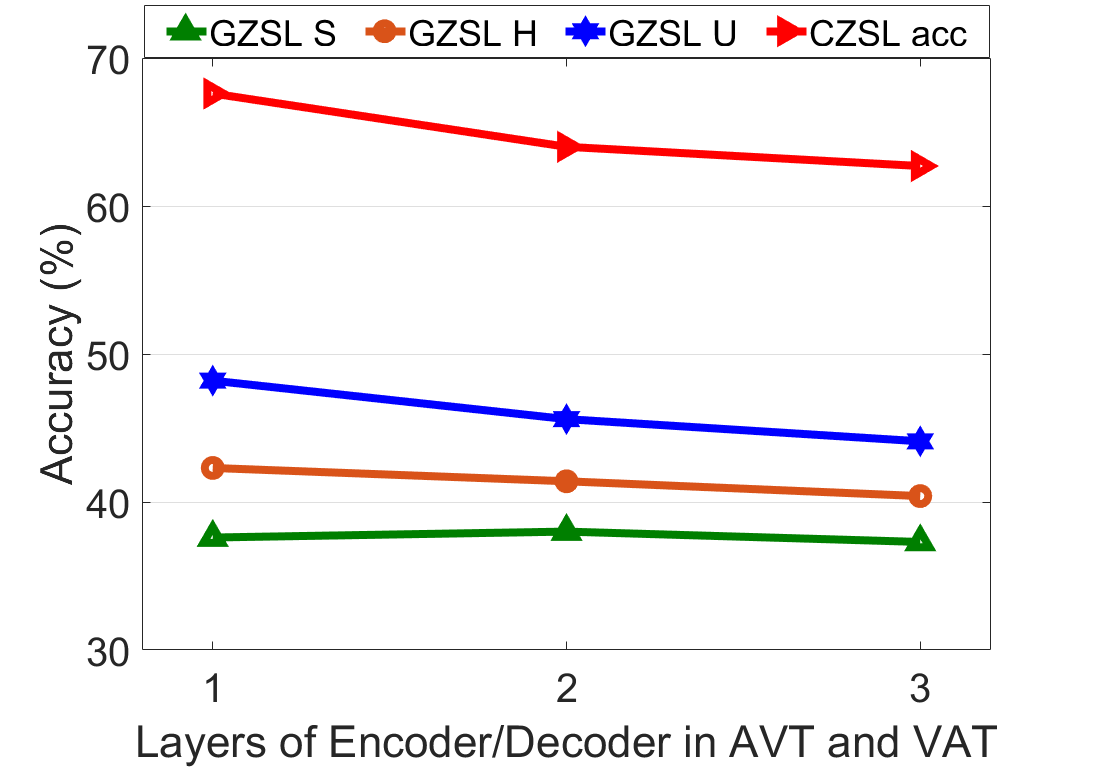}
			\includegraphics[width=4.1cm,height=3.2cm]{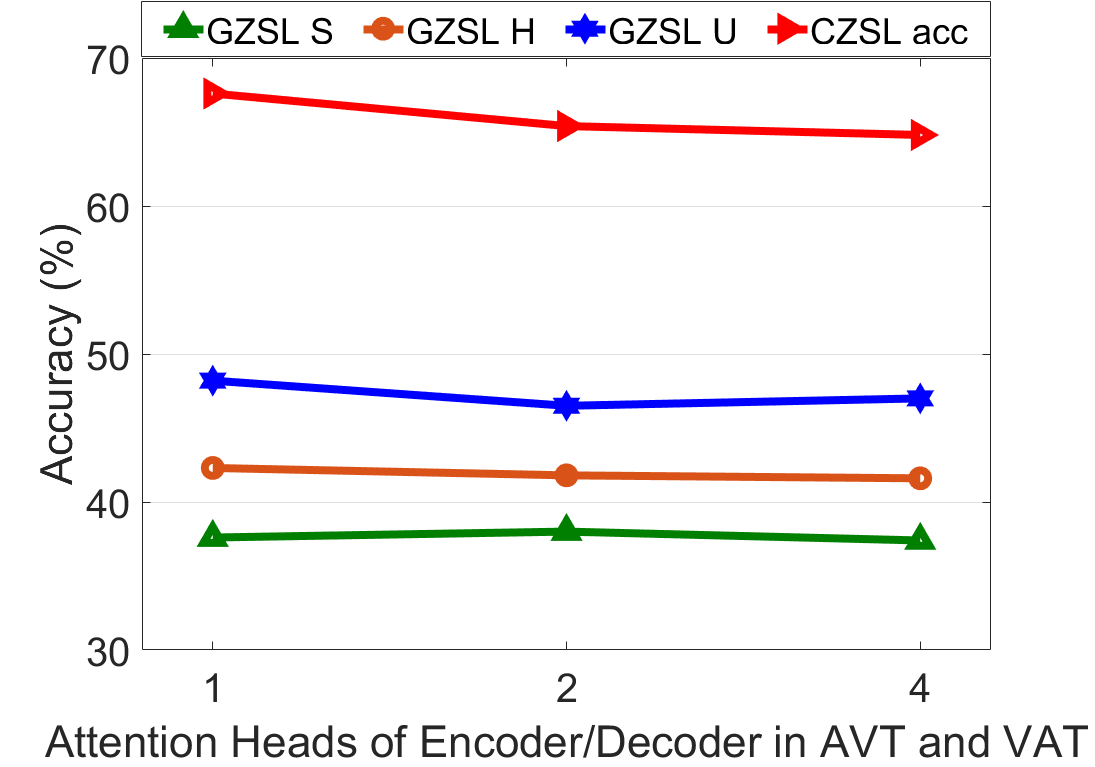}\\
			\caption{The effects of different architectures for the AVT and VAT networks on CUB and SUN. We investigate the number
				of layers of Encoder/Decoder and attention heads in Encoder/Decoder.}
			\label{fig:architecture}\vspace{-6mm}
		\end{center}
	\end{figure}

	\begin{figure}[t]
		\begin{center}
			\includegraphics[width=9.3cm,height=4.6cm]{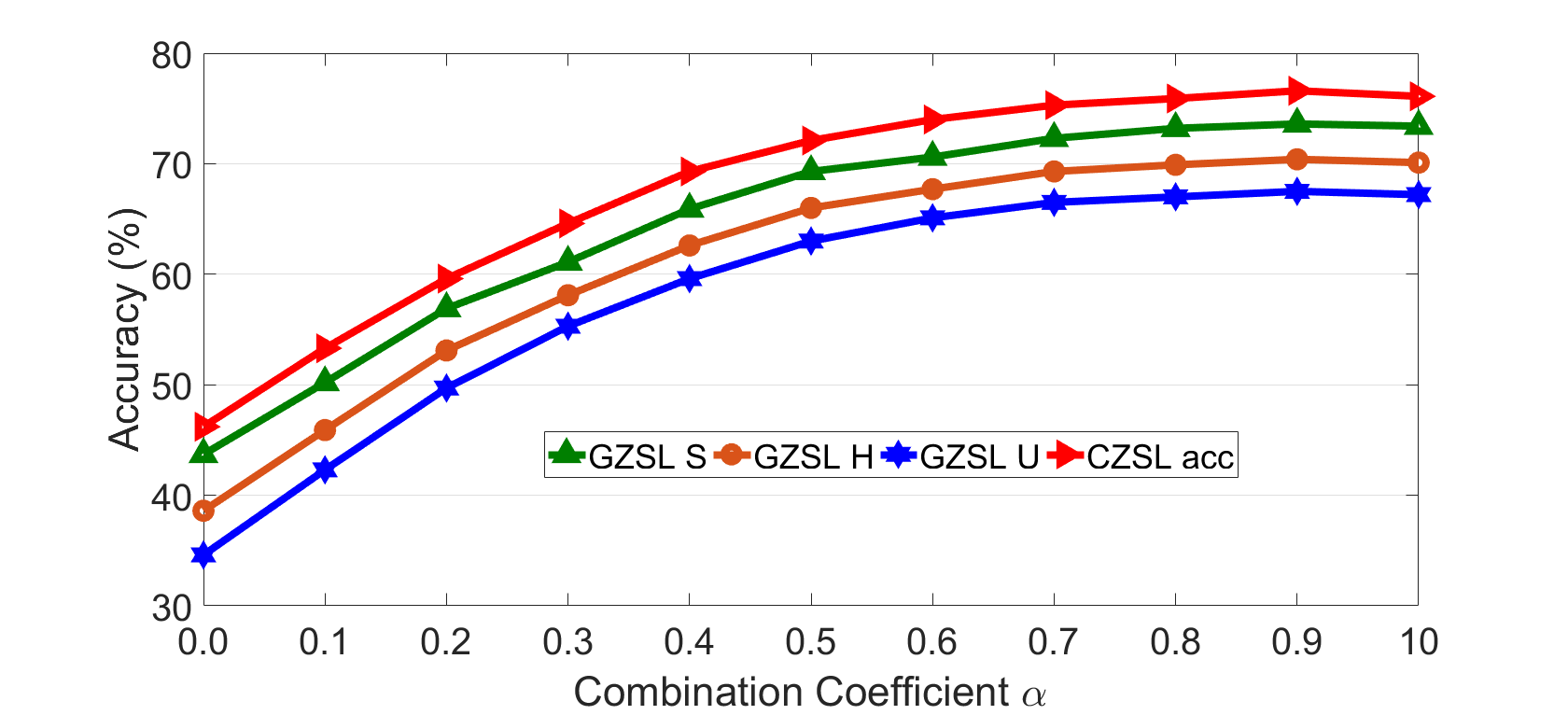}\vspace{2mm}
			\includegraphics[width=9.3cm,height=4.6cm]{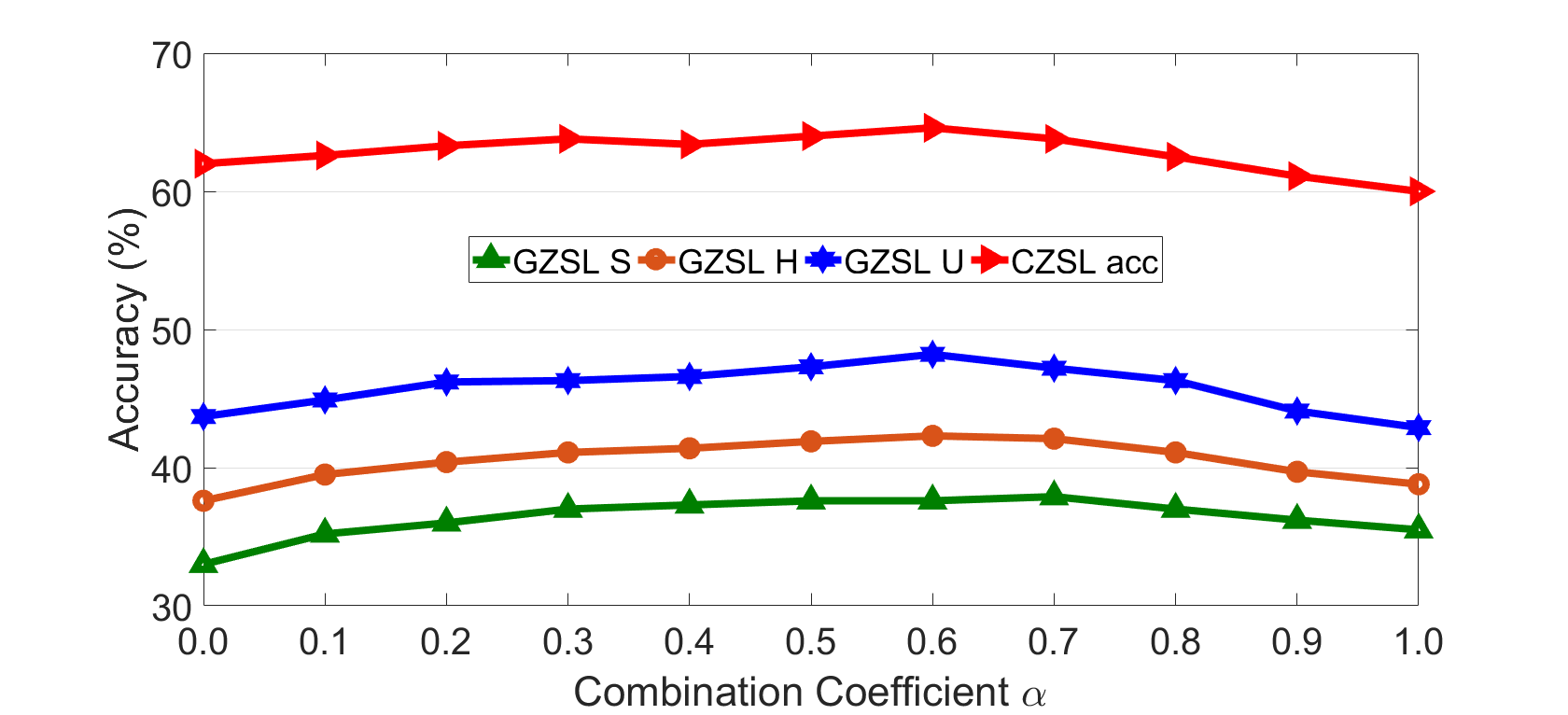}\\
			\caption{The effectiveness of combination coefficients $\alpha$ between the AVT and VAT sub-nets on CUB (top) and SUN (bottom).}
			\label{fig:combination}\vspace{-6mm}	
		\end{center}
	\end{figure}
	
	\subsection{Part Localization Prediction}\label{sec4.4}	
	Together with attention map visualizations, we quantitatively report the part localization prediction by calculating the Percentage of Correctly Localized Parts (PCP) of unseen classes on CUB. Our calculation follows that in APN\footnote{https://github.com/wenjiaXu/APN-ZSL} \cite{Xu2020AttributePN}. When the predicted bounding box for a part overlaps sufficiently with the grounding truth bounding box (i.e., IoU $>$ 0.5), the detection is considered to be correct. Table \ref{table:PCP} shows the results where our TransZero++ consistently and significantly improves the localization accuracy of all parts over DAZLE \cite{Huynh2020FineGrainedGZ} and APN \cite{Xu2020AttributePN}. These results are in accordance to the qualitative results in Fig. \ref{fig:attened-part} where our TransZero++ can better discover the key semantic knowledge between visual and attribute features, resulting in desirable knowledge transfer for ZSL. 
	\begin{table}[h]
		\renewcommand{\thetable}{6}
		\centering
		\caption{Results of part localization prediction for various methods on CUB. For BB (bounding box) size, $1/\sqrt{2}$ denotes each part
			bounding box has the size of $1/\sqrt{2}W\times1/\sqrt{2}H$, where $W$ and $H$ are the width and height of the bird image.} \label{table:PCP}
		\vspace{-2mm}
		\resizebox{1.0\linewidth}{!}{
			{
				\begin{tabular}{l|c|cccccc|c}
					\hline
					Methods& BB size &Head&Breast&Belly&Back&Wing&Leg& Mean\\
					\hline
					DAZLE [35]        &$1/\sqrt{2}$&94.7&92.1&75.1&	65.1&68.4&	50.5&74.3\\
					APN [39] &$1/\sqrt{2}$&91.8&88.9&81.0&72.1&	76.6&65.0&79.2\\
					TransZero++                                 &$1/\sqrt{2}$&97.9&93.7&	85.3&80.4&86.0&	74.0&86.2\\
					\hline
			\end{tabular}}
		}\vspace{-4mm}
	\end{table}

	\subsection{Generality Analysis}\label{sec4.5}
	To show the generality of our TransZero++, we conduct experiments by replacing the attributes descriptions with word vectors of classes name (denotes w2v) \cite{Mikolov2013DistributedRO}. Results are shown in Table \ref{table:w2v}. Since the attribute descriptions provide more informative representations than w2v, all ZSL methods achieve inferior performance using w2v compared to attributes. Fortunately, we observe that our method achieves the best result of 25.1\% and 28.7\% in the CZSL setting on CUB and SUN, respectively. In the GZSL setting, TransZero++ performs significant gains of harmonic mean with 5.1\% over APN \cite{Xu2020AttributePN} on SUN dataset. These results indicate that TransZero++ can also discover the key semantic knowledge with the guidance of w2v. This further demonstrates the advantages of our method.
	\begin{table}[h]
		\centering
		\caption{ Results of various methods with word vector of class names (denotes as w2v). “$\dagger$” indicates that results are taken from \cite{Xu2022VGSEVS}.} \label{table:w2v}
		\vspace{-2mm}
		\resizebox{1.0\linewidth}{!}{
			{
				\begin{tabular}{l|c|ccc|c|ccc}
					\hline
					\multirow{2}*{Method} &\multicolumn{4}{c|}{CUB} &\multicolumn{4}{c}{SUN}\\
					\cline{2-5}\cline{6-9}
					&\rm{acc}&\rm{U} & \rm{S} & \rm{H} &\rm{acc}&\rm{U} & \rm{S} & \rm{H}\\
					\hline
					\rowcolor{mygray}SJE$^\dagger$ \cite{Akata2015EvaluationOO}    &14.4&13.2 &28.6& 18.0 &26.3&19.8& 18.6& 19.2\\
					APN$^\dagger$ \cite{Xu2020AttributePN}      &22.7& 17.6&29.4&22.1&23.6&16.3&15.3&15.8\\
					\rowcolor{mygray}TransZero++        &25.1&12.9&29.9&18.0&28.7&16.4&28.7&20.9\\
					\hline
			\end{tabular}}
		}\vspace{-4mm}
	\end{table}
	
	\begin{table}[t]
		\centering
		\caption{ Results of various generative models with visual features extracted from TransZero++.} \label{table:generative}
		\vspace{-2mm}
		\resizebox{1.0\linewidth}{!}{
			{
				\begin{tabular}{l|c|ccc|c|ccc}
					\hline
					\multirow{2}*{Method} &\multicolumn{4}{c|}{CUB} &\multicolumn{4}{c}{SUN}\\
					\cline{2-5}\cline{6-9}
					&\rm{acc}&\rm{U} & \rm{S} & \rm{H} &\rm{acc}&\rm{U} & \rm{S} & \rm{H}\\
					\hline
					\rowcolor{mygray}f-VAEGAN~\cite{Xian2019FVAEGAND2AF}&61.0&48.4&60.1& 53.6&64.7&45.1&38.0&41.3\\
					TransZero++ $\oplus$ f-VAEGAN~\cite{Xian2019FVAEGAND2AF}&71.2&	61.7& 59.7&60.7&66.1&50.4&35.5&41.6\\
					\rowcolor{mygray}TF-VAEGAN~\cite{Narayan2020LatentEF}& 64.9&52.8&64.7&58.1&66.0&45.6&40.7& 43.0 \\
					TransZero++ $\oplus$ TF-VAEGAN~\cite{Narayan2020LatentEF}        &71.5&63.3& 62.8& 63.1&68.5&48.3& 40.9&44.3\\
					\hline
			\end{tabular}}
		}\vspace{-3mm}
	\end{table}
	
	Furthermore, we also extract feature representations from the decoders of our TransZero++ and entailing them on the top of generative models, \textit{e.g.}, f-VAEGAN\cite{Xian2019FVAEGAND2AF} and TF-VAEGAN \cite{Narayan2020LatentEF}. Results in Table \ref{table:generative} show that our TransZero++ consistently boosts the performances of generative models on two datasets. Specifically, TransZero++ improves the performance of f-VAEGAN on $\bm{acc}/\bm{H}$ with 10.2\%/7.1\% and 1.4\%/0.3\% on CUB and SUN, respectively. TransZero++ also gains the improvements of TF-VAEGAN on $\bm{acc}/\bm{H}$ with  6.6\%/5.0\% and 2.5\%/1.3\% on CUB and SUN, respectively. These results indicate that our TransZero++ discovers the key semantic knowledge to represent the semantic-augmented features, helping the generative models synthesize discriminative and transferable visual features. As such, the cross-dataset bias problem \cite{Chen2021FREE} in f-VAEGAN and TF-VAEGAN can be alleviated.

	Finally, we also extend TransZero++ to the transductive ZSL setting following \cite{Song2018TransductiveUE}, where unlabeled samples of unseen classes are also used for model optimization. We compare TransZero++ with other embedding-based Transductive ZSL methods on CUB and SUN datasets, as shown in Table \ref{table:transductive}. Compared to other embedding-based transductive ZSL methods, our TransZero++ achieves new state-of-the-art of $\bm{acc}/\bm{H}$ with 81.5\%/77.7\% and 69.0\%/48.8\% on CUB and SUN, respectively. This indicates that TransZero++ also learns the intrinsic semantic knowledge for desirable knowledge transfer in transductive ZSL.
	
	\begin{table}[h]
		\centering
		\caption{ Results of various embedding-based transductive ZSL methods on CUB and SUN datasets.} \label{table:transductive}
		\vspace{-2mm}
		\resizebox{1.0\linewidth}{!}{
			{
				\begin{tabular}{l|c|ccc|c|ccc}
					\hline
					\multirow{2}*{Method} &\multicolumn{4}{c|}{CUB} &\multicolumn{4}{c}{SUN}\\
					\cline{2-5}\cline{6-9}
					&\rm{acc}&\rm{U} & \rm{S} & \rm{H} &\rm{acc}&\rm{U} & \rm{S} & \rm{H}\\
					\hline
					\rowcolor{mygray}ALE-tran~\cite{Xian2019ZeroShotLC}&54.5&23.5&45.1&30.9& 55.7&19.9&22.6&21.2\\
					DSRL~\cite{Ye2017ZeroShotCW}& 48.7&17.3&39.0& 24.0&56.8&17.7& 25.0& 20.7 \\
					\rowcolor{mygray}UE-finetune~\cite{Song2018TransductiveUE}&72.1&74.9&71.5&73.2&58.3&33.6&54.8&41.7\\
					TransZero++       &81.5&79.6&75.8&77.7&69.0&64.9&39.1&48.8\\
					\hline
			\end{tabular}}
		}\vspace{-3mm}
	\end{table}

	\subsection{Hyperparameter Analysis}\label{sec4.6}	
	To analyse the robustness of our TransZero++ and select better hyperparameters for it. We conduct extensive experiments for evaluating the effects of loss weights (in Eq. \ref{Eq:L_AVT} and Eq. \ref{Eq:L_final}), Transformer architecture settings in AVT and VAT, and combination coefficient (in Eq. \ref{eq:prediction}). 
	
	\subsubsection{Effects of Loss Weights}\label{4.4.1}
	Here, we analyse the effects of loss weights that control their corresponding loss terms, \textit{i.e.},  $\lambda_{\text{AR}}$, $\lambda_{\text{SC}}$, $\lambda_{\text{VAT}}$, $\lambda_{\text{f\_SCL}}$ and $\lambda_{\text{p\_SCL}}$. We try a range of these loss weights evaluated on CUB and SUN, \textit{i.e.}, $\{0.0001, 0.001, 0.01, 0.1, 1.0\}$. Results are shown in Fig. \ref{fig:loss-weights}. When $\lambda_{AR}$, $\lambda_{\text{f\_SCL}}$ and $\lambda_{\text{p\_SCL}}$ are set to a large value, all evaluation protocols tend to drop. Moreover, TansZero++ are relatively insensitive to $\lambda_{\text{SC}}$ and $\lambda_{VAT}$ when they are set to small (\textit{e.g.}, smaller than 0.01).  Based on these observations, we set $\{\lambda_{\text{AR}}, \lambda_{\text{SC}}, \lambda_{\text{VAT}}, \lambda_{\text{f\_SCL}},\lambda_{\text{p\_SCL}}\}$ to $\{0.0001, 0.1, 0.1, 0.001, 0.01\}$ and $\{0.01, 0.1, 1.0, 0.001,\\0.001\}$ for CUB and SUN datasets, repectively.
	
	\subsubsection{Effects of Different Architectures for Transformer in AVT and VAT}\label{4.4.3} 
	To find the best Transformer settings in AVT and VAT, we investigate the influence of the number of i) layers of Encoder/Decoder, and ii) attention heads of Encoder/Decoder. To enable the training of TrasZero++ to be more stable, we set same number of layers/heads in the encoder and decoder. As shown in Fig. \ref{fig:architecture}, we find that the encoders/decoders in AVT and VAT should be set to be small, \textit{i.e.}, one layer with one attention head, TransZero++ can achieve better results in both seen and unseen classes. The possible reason lies in that the training data for the ZSL model is medium/small scale which inevitably leads to over-fitting with more complex Transformer architectures.
	
	\subsubsection{Effects of Combination Coefficient}\label{4.4.2} 
	We argue that the attribute-based visual features and visual-based attribute features learned by AVT and VAT respectively are complementary, and thus we take a combination coefficient $\alpha$ to fuse their corresponding semantic-augmented embeddings for desirable visual-semantic interaction (in Eq. \ref{eq:prediction}). We try a range of $\alpha$ on CUB and SUN, \textit{i.e.}, $\{0.0,0.1, 0.2, 0.3, 0.4, 0.5, 0.6, 0.7, 0.8, 0.9, 1.0\}$. Notably, $\alpha=0.0$ is denoted as the TransZero++(VAT), and $\alpha=1.0$ is denoted as the TransZero++(AVT). As shown in Fig. \ref{fig:combination}, when $\alpha$ is set to large relatively (\textit{e.g.}, $\alpha>0.5$), TransZero++ achieves better results. This demonstrates that AVT sub-net provides more desirable information for TransZero++. However, $\alpha$ should also not be set too large since the VAT sun-net provides additional useful information for TransZero++. Based on these results, we set $\alpha$ to 0.9 and 0.6 for CUB and SUN, respectively.

	\section{Conclusion}\label{sec5}
	In this paper, we propose a novel cross attribute-guided Transformer network for ZSL, termed TransZero++. TransZero++ consists of a attribute$\rightarrow$visual Transformer sub-net (AVT) and visual$\rightarrow$attribute Transformer sub-net (VAT). First, AVT employs a feature augmentation encoder to improve the discriminability and transferability of visual features by alleviating the cross-dataset problem and reducing the entangled region feature relationships, respectively. Meanwhile, an attribute$\rightarrow$visual decoder in AVT is introduced to learn the attribute localization for attribute-based visual feature representations which are locality-augmented.  Secondly, VAT applied a similar feature augmentation encoder to refine the visual features, which is further fed into a visual$\rightarrow$attribute decoder to learn the visual-based attribute features. By introducing the feature-level and prediction-level semantical collaborative losses for optimization, our TransZero++ can learn the semantic-augmented visual embedding. Considering the attribute-based visual features and visual-based attribute features that are complementary to each other, we combine the two semantic-augmented visual embeddings learned by AVT and VAT to enable desirable visual-semantic interaction cooperated with the class semantic vectors for ZSL classification. Extensive experiments on three popular ZSL benchmarks and on the large-scale ImageNet dataset demonstrate the superiority of our method. We believe that our work also facilitates the development of other visual-and-language learning systems, \textit{e.g.}, image captioning \cite{Herdade2019ImageCT}, natural language for visual reasoning \cite{Antol2015VQAVQ}.
	
	Indeed, our TransZero++ bases on the Glove embedding of attribute names, which is not easy to get in real-life. Since w2v lacks enough informative representations, it cannot well supports ZSL methods to significantly discover the key semantic knowledge \cite{Xu2022VGSEVS,Fu2018RecentAI}. Motivated by the work of Xu \cite{Xu2022VGSEVS}, we can effectively discover semantic embeddings containing discriminative visual properties via visually clustering and class relations prediction, without requiring any human annotation to replace the attributes defined by experts in future work.

	\section*{Acknowledgements}  This work is partially supported by NSFC~(62172177, 62006244,61929104,62276134), National Key R\&D Program (2022YFC3301004,2022YFC3301704), NSF of Hubei Province (2021CFB332), and Young Elite Scientist Sponsorship Program of China Association for Science and Technology (YESS20200140).

	\ifCLASSOPTIONcaptionsoff
	\newpage
	\fi
	
	\bibliographystyle{IEEEtran}
	\bibliography{mybibfile}
	
	\newpage
	\vspace{-1.1cm}
	\begin{IEEEbiography}[{\includegraphics[width=1in,height=1.25in,clip,keepaspectratio]{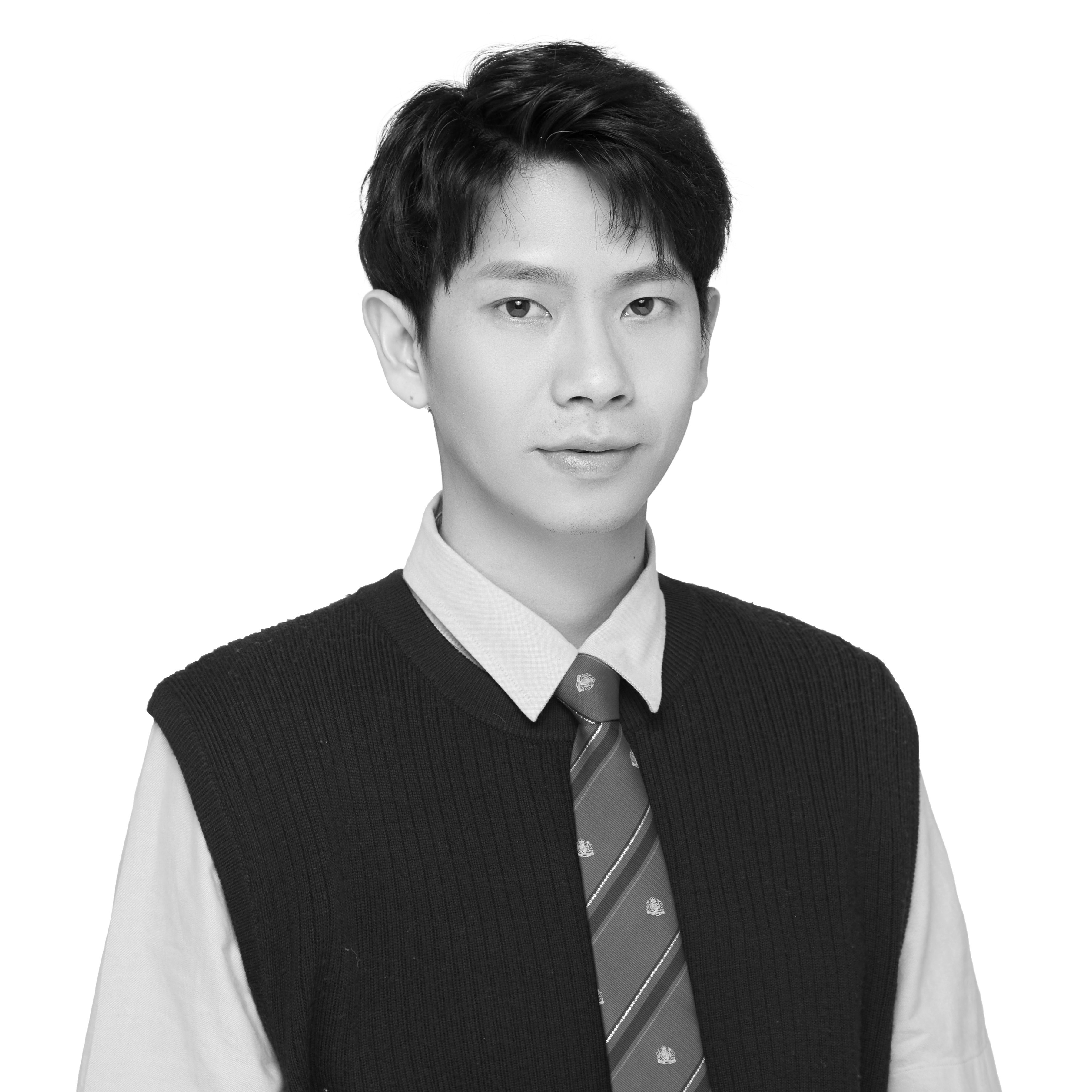}}]{Shiming Chen}
		is currently a full-time Ph.D. student in the School of Electronic Information and Communications, Huazhong University of Sciences and Technology (HUST), China. His research results have expounded in prominent conferences and prestigious journals, such as NeurIPS, ICCV, CVPR, AAAI, IJCAI, IEEE TNNLS, IEEE TEVC, and etc. His current research interests span computer vision and machine learning with a series of topics, such as generative modeling and learning, zero-shot learning, and visual-and-language learning. 
	\end{IEEEbiography}

	\vspace{-1.13cm}
	\begin{IEEEbiography}[{\includegraphics[width=1in,height=1.25in,clip,keepaspectratio]{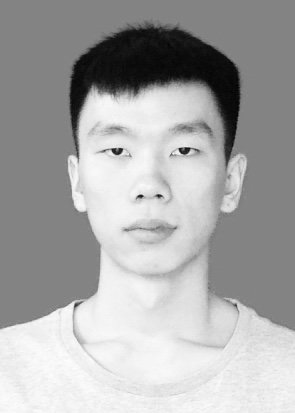}}]{Ziming Hong}
		is currently pursuing the M.Sc. degree in the School of Electronic Information and Communications(EIC), Huazhong University of Sciences and Technology(HUST), China. He received the B.E. degree in the School of Information Engineering, Wuhan University of Technology(WHUT), in 2019. His current research interests include graph learning and zero-shot learning.
	\end{IEEEbiography}
	
	\vspace{-1.2cm}
	\begin{IEEEbiography}[{\includegraphics[width=1in,height=1.25in,clip,keepaspectratio]{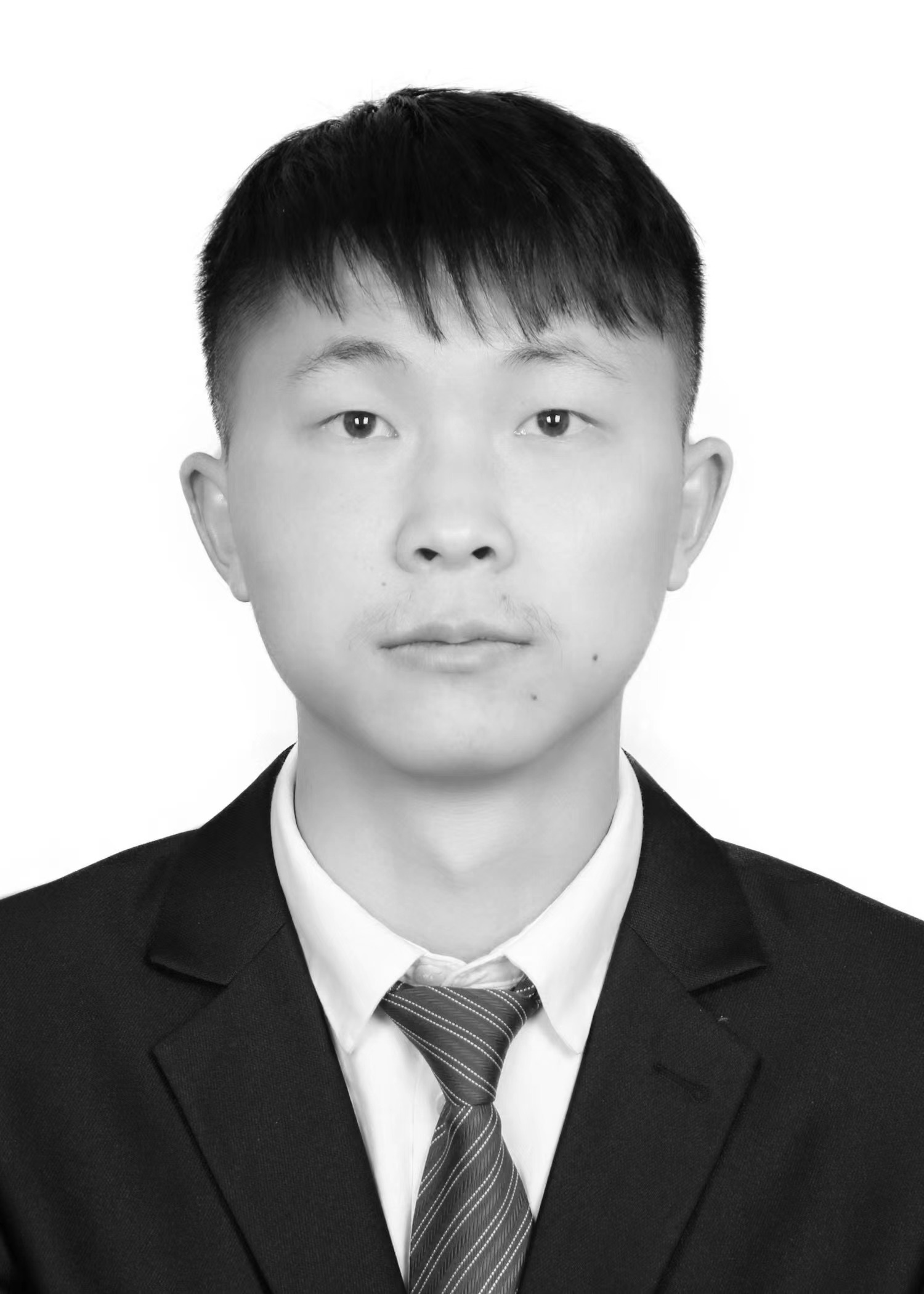}}]{Wenjin Hou}
		is currently a full-time M.Sc. student at the School of Electronic Information and Communication, Huazhong University of Science and Technology (HUST), China. He received the B.E. degree in the School of Information Science and Engineering, Lanzhou University (LZU), in 2021. His research interests include zero-shot learning and generative modeling and learning in the field of computer vision.
	\end{IEEEbiography}

	\vspace{-1.2cm}
	\begin{IEEEbiography}[{\includegraphics[width=1in,height=1.25in,clip,keepaspectratio]{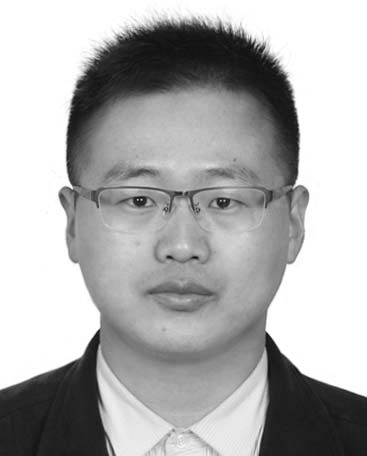}}]{Guo-Sen Xie}
		is currently a Professor at the School of Computer Science and
		Engineering, Nanjing University of Science and Technology, China. He received his Ph.D. degree from the National Laboratory of Pattern Recognition, Institute of Automation, Chinese Academy of Sciences, Beijing, China, in 2016. His research interests include computer vision and machine learning.
	\end{IEEEbiography}
	
	\begin{IEEEbiography}
		[{\includegraphics[width=1in,height=1.25in,clip,keepaspectratio]{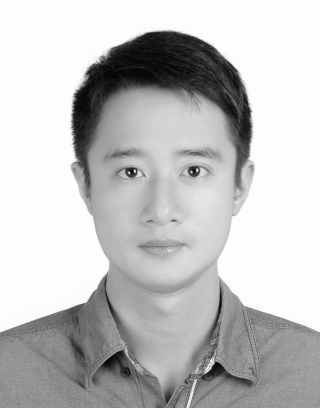}}]
		{Yibing Song} is with AI$^3$ Institute, Fudan University. He was a senior researcher at Tencent AI Lab. He has obtained a Ph.D. degree from City University of Hong Kong, a MPhil degree from the same university, and a bachelor degree from University of Science and Technology of China. He has served as area chairs for CVPR, NeurIPS, and ICLR, served as reviewers for premier computer vision and machine learning conferences, and received multiple outstanding reviewer awards in CVPR 2018-2020, ECCV 2022, and NeurIPS 2019.
	\end{IEEEbiography}

	\vspace{-1cm}
	\begin{IEEEbiography}[{\includegraphics[width=1in,height=1.25in,clip,keepaspectratio]{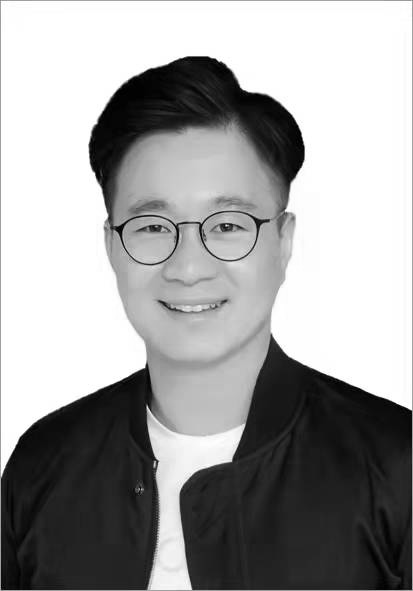}}]{Jian Zhao}
		is currently an Assistant Professor with the Institute of North Electronic Equipment, Beijing, China. He received his Ph.D. degree from the National University of Singapore (NUS) in 2019. He has served as the guest editor of PRL and Electronics, the presentation chair of the CICAI'21, the session chair of the ACM MM'21, and the invited reviewer of NSFC, T-PAMI, IJCV, NeurIPS, CVPR, etc. He has received the "2020-2022 Young Elite Scientist Sponsorship Program" from China Association for Science and Technology, and the "2021-2023 Beijing Young Elite Scientist Sponsorship Program" from Beijing Association for Science and Technology. His main research interests include deep learning, pattern recognition, computer vision and multimedia. He has published over 40 cutting-edge papers (e.g., T-PAMI, IJCV, NeurIPS, CVPR, etc.). He has received the nomination for the USERN Prize 2021, and won the Lee Hwee Kuan Award (Gold Award) on PREMIA'19 and the "Best Student Paper Award" on ACM MM'18 as the first author. 
	\end{IEEEbiography}

	\vspace{-0.6cm}
	\begin{IEEEbiography}[{\includegraphics[width=1in,height=1.25in,clip,keepaspectratio]{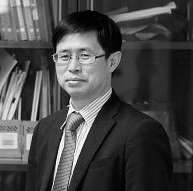}}]{Xinge You}
		(Senior Member, IEEE) is currently a Professor with the School of Electronic Information and Communications, Huazhong University of Science and Technology, Wuhan. He received the B.S. and M.S. degrees in mathematics from Hubei University, Wuhan, China, in 1990 and 2000, respectively, and the Ph.D. degree from the Department of Computer Science, Hong Kong Baptist University, Hong Kong, in 2004. His research results have expounded in 60+ publications at prestigious journals and prominent conferences, such as  IEEE T-PAMI, T-IP, T-NNLS, NeurIPS, CVPR, ICCV, ECCV, and etc. He served/serves as an Associate Editor of the \textit{IEEE Transactions on Cybernetics}, \textit{IEEE Transactions on Systems, Man, Cybernetics:Systems}. His current research interests include image processing, wavelet analysis and its applications, pattern recognition, machine earning, and computer vision.
	\end{IEEEbiography}

	\vspace{-0.6cm}
	\begin{IEEEbiography}[{\includegraphics[width=1in,height=1.25in,clip,keepaspectratio]{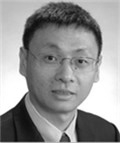}}]{Shuicheng Yan}
		(Fellow, IEEE) is currently the director of Sea AI Lab (SAIL) and group chief scientist of Sea. He is an Fellow of Academy of Engineering, Singapore, IEEE Fellow, ACM Fellow, IAPR Fellow. His research areas include computer vision, machine learning and multimedia analysis. Till now, he has published over 600 papers in top international journals and conferences, with Google Scholar Citation over 40,000 times and H-index 105. He had been among “Thomson Reuters Highly Cited Researchers” in 2014, 2015, 2016, 2018, 2019. Dr. Yan’s team
		has received winner or honorable-mention prizes for 10 times of two core competitions, Pascal VOC and ImageNet (ILSVRC), which are deemed as “World Cup” in the computer vision community. Also his team won over 10 best paper or best student paper prizes and especially, a grand slam in ACM MM, the top conference in multimedia, including Best Paper Award, Best Student Paper Award and Best Demo Award.
	\end{IEEEbiography}

	\vspace{-0.6cm}
	\begin{IEEEbiography}[{\includegraphics[width=1in,height=1.25in,clip,keepaspectratio]{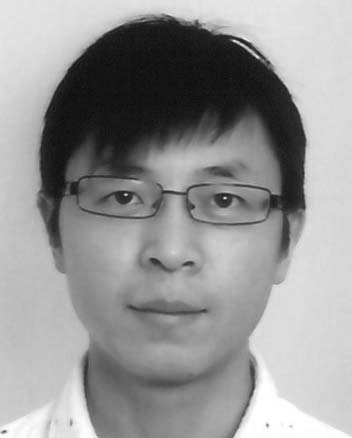}}]{Ling Shao}
		(Fellow, IEEE)   is the Chief Scientist of Terminus Group and the President of Terminus International. He was the founding CEO and Chief Scientist of the Inception Institute of Artificial Intelligence, Abu Dhabi, UAE. His research interests include computer vision, deep learning, medical imaging and vision and language. He is a fellow of the IEEE, the IAPR, the BCS and the IET. 
	\end{IEEEbiography}

\end{document}